\documentclass[10pt,journal,compsoc]{IEEEtran}

\ifCLASSOPTIONcompsoc
  \usepackage[nocompress]{cite}
\else
  \usepackage{cite}
\fi

\ifCLASSINFOpdf
  \usepackage[pdftex]{graphicx}
\else
\fi

\usepackage{amsmath}

\ifCLASSOPTIONcompsoc
 \usepackage[caption=false,font=footnotesize,labelfont=sf,textfont=sf]{subfig}
\else
 \usepackage[caption=false,font=footnotesize]{subfig}
\fi

\usepackage{stfloats}

\usepackage{url}

\usepackage{amssymb}

\usepackage{booktabs}       %
\usepackage{threeparttable}
\usepackage[hypcap=false]{caption}
\usepackage[table]{xcolor}         %
\usepackage{colortbl} %
\usepackage{multirow}
\usepackage{enumitem}
\definecolor{tabletitle}{gray}{.8}
\definecolor{ours}{gray}{.95}
\definecolor{ggray}{RGB}{127,127,127}
\definecolor{reda}{RGB}{202,0,0}
\definecolor{redb}{RGB}{217,148,143}
\definecolor{myyellow}{RGB}{190,144,0}
\definecolor{mygreen}{RGB}{0,136,51}
\definecolor{myblue}{RGB}{0,102,204}
\newcolumntype{B}{!{\vrule width 1pt}}
\usepackage{pifont}
\newcommand{\yes}{\text{\ding{51}}}  %
\newcommand{\blank}{—}

\newcommand{\best}[1]{\textcolor{reda}{\textbf{#1}}}
\newcommand{\second}[1]{\textcolor{mygreen}{\textbf{#1}}}
\newcommand{\third}[1]{\textcolor{myblue}{\textbf{#1}}}

\usepackage{algorithm,algorithmicx,algpseudocode}
\algrenewcommand\algorithmicindent{0.5em}%

\usepackage[pagebackref=true,breaklinks=true,letterpaper=true,colorlinks,bookmarks=false]{hyperref}

\begin{document}
\title{ComPtr: Towards Diverse Bi-source Dense Prediction Tasks via A Simple yet General Complementary Transformer}

\author{Youwei~Pang,~
  Xiaoqi~Zhao,~
  Lihe~Zhang,~
  and~Huchuan~Lu,~\IEEEmembership{Fellow,~IEEE}%
  \IEEEcompsocitemizethanks{
    \IEEEcompsocthanksitem All the authors are with Dalian University of Technology, Dalian, China.
    \IEEEcompsocthanksitem Y. Pang: \url{lartpang@gmail.com}.
    \IEEEcompsocthanksitem L. Zhang is the corresponding author.
  }%
  \thanks{Preprint version.}
}

\IEEEtitleabstractindextext{%
  \begin{abstract}
    Deep learning (DL) has advanced the field of dense prediction, while gradually dissolving the inherent barriers between different tasks.
    However, most existing works focus on designing architectures and constructing visual cues only for the specific task, which ignores the potential uniformity introduced by the DL paradigm.
    In this paper, we attempt to construct a novel \underline{ComP}lementary \underline{tr}ansformer, \textbf{ComPtr}, for diverse bi-source dense prediction tasks.
    Specifically, unlike existing methods that over-specialize in a single task or a subset of tasks, ComPtr starts from the more general concept of bi-source dense prediction.
    Based on the basic dependence on information complementarity, we propose consistency enhancement and difference awareness components with which ComPtr can evacuate and collect important visual semantic cues from different image sources for diverse tasks, respectively.
    ComPtr treats different inputs equally and builds an efficient dense interaction model in the form of sequence-to-sequence on top of the transformer.
    This task-generic design provides a smooth foundation for constructing the unified model that can simultaneously deal with various bi-source information.
    In extensive experiments across several representative vision tasks, i.e. remote sensing change detection, RGB-T crowd counting, RGB-D/T salient object detection, and RGB-D semantic segmentation, the proposed method consistently obtains favorable performance.
    The code will be available at \url{https://github.com/lartpang/ComPtr}.
  \end{abstract}

  }

\maketitle

\IEEEdisplaynontitleabstractindextext

\IEEEraisesectionheading{\section{Introduction}\label{sec:introduction}}

\IEEEPARstart{W}{ith} the development of vision transformer~\cite{ViT,Swin}, recent works achieve stunning performance on many important benchmarks.
The task-generic characteristic of the sequence-to-sequence interaction architecture prompts the gaps among different tasks to be further eliminated.

Dense prediction is an important and fundamental problem of computer vision.
It often requires laborious and dense annotations of various objects of interest.
Typical tasks include
remote sensing image segmentation~\cite{changedetection-LEVIR,changedetection-SYSU},
crowd counting (density map estimation)~\cite{rgbtcc-CMCRL-RGBTCC},
salient object detection~\cite{RGBDSODSurveyZhou,CMMSODSurvey,WWGSODSurvey},
and scene image semantic segmentation~\cite{SUN-RGBD-37classes}.
Existing schemes have achieved impressive performance in their respective tasks.
Among them, the bi-source\footnote{We use ``bi-source'' to emphasize the form of the data source of interest, which in effect encompasses both single-modal (bi-temporal) and bi-modal input forms as shown in Tab.~\ref{tab:task}.} input paradigm is a common style due to its practicability and effectiveness.
Visual information from different sources can provide a more comprehensive context for scene semantics geared towards specific task requirements.
For example, the exploration of scene monitoring in a large time span~\cite{changedetection-LEVIR,changedetection-SYSU} and scene understanding in complex environments~\cite{rgbtcc-CMCRL-RGBTCC,SIP,VT5000-ADF,SUN-RGBD-37classes} rely on more auxiliary data corresponding to different times and different modalities for joint information mining, respectively.
By integrating different sources, it would be easier to perceive objects and contexts.
However, because of the fragmented definitions, these approaches usually focus on a single task.
This task-specific design brings repetitive exploration, redundant parameters, and resource overhead.
In this work, we explore the potential commonalities of architecture design and visual semantics among various tasks.

\begin{figure}[!t]
  \centering
  \includegraphics[width=0.8\linewidth]{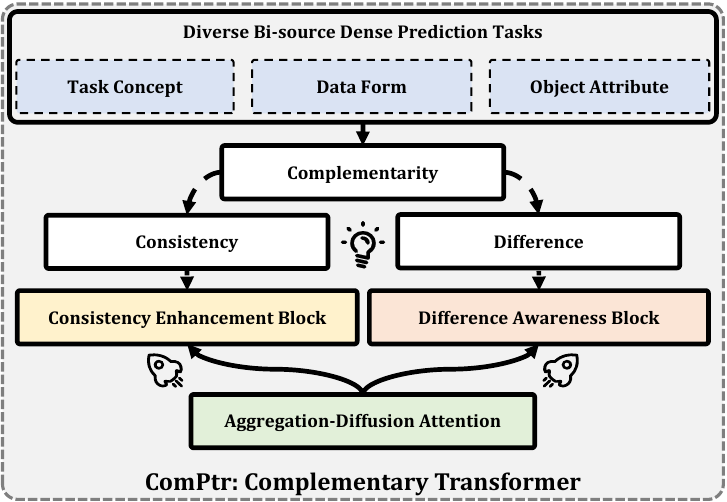}
  \setlength{\abovecaptionskip}{0.5em}
  \caption{Visual outline of the proposed ComPtr.
    For diverse bi-source dense prediction tasks, our method is based on the generic concept of complementarity.
    The consistency enhancement block (Sec.~\ref{sec:consistencyblock}) and difference awareness block (Sec.~\ref{sec:differenceblock}) are constructed based on the proposed efficient aggregation-diffusion attention (Sec.~\ref{sec:aggdffattn}) from the perspectives of consistency and difference, respectively.
    Such a design as a whole serves the implementation of the task-generic architecture.
  }
  \label{fig:teaser}
  \vspace{-1em}
\end{figure}

{Existing methods~\cite{rgbdtsod-SwinNet,caver-tip,rgbdtsod-HRTransNet,mmseg-DFormer,mmseg-CMX,mmseg-TokenFusion} often directly build cross-modal dense interactions to implicitly explore inter-modal correlations.
  {Different from them}, we design a top-down heuristic modeling strategy and starts from the basic needs of bi-source dense prediction tasks, {\em i.e.}, the complementarity between two inputs.
We explicitly decompose it into consistency and difference representation modeling as depicted in Fig.~\ref{fig:teaser}.
This design well accommodates the diverse modeling needs of tasks in Tab.~\ref{tab:task}.}
Based on it, we design a simple yet effective framework ComPtr, as shown in Fig.~\ref{fig:net}, to address the general bi-source vision modeling.
Specifically, we design a complementary transformer for diverse bi-source vision tasks, which contains consistency enhancement blocks and difference awareness blocks, {\em i.e.}, CEM and DAM.
The ``consistency'' tends to highlight the commonality between multi-source data.
The CEM blocks are equipped in the encoder to strengthen the object-related representation.
While the ``difference'' depicts the specificity of each stream.
We build the DAM-based decoder, which continuously endows the enhanced common representation with scene details from different sources to achieve accurate predictions.

\begin{table*}[!t]
  \centering
  \resizebox{0.75\linewidth}{!}{\begin{tabular}{c|cc|ccc|cc}
  \toprule[2pt]
                                                 &
  \multicolumn{2}{c}{Input Attributes}           &
  \multicolumn{5}{|c}{Output Attributes}                                                                                                                                                   \\
  \cline{2-8}
                                                 &
                                                 &
                                                 &
  \multicolumn{3}{c}{Object Category Attributes} &
  \multicolumn{2}{|c}{Task Requirements}                                                                                                                                                   \\
  \cline{4-8}
  \multirow{-3}{*}{Selected Tasks}               & \multirow{-2}{*}{Single-Modal} & \multirow{-2}{*}{Multi-Modal} & Class-agnostic & Single-class & Multi-class & Region Seg. & Point Loc. \\
  \midrule[1pt]
  Remote Sensing Change Detection (RSCD)                & \yes                           &                               &                & \yes         &             & \yes        &            \\
  RGB-T Crowd Counting (RGB-T CC)                           &                                & \yes                          &                & \yes         &             &             & \yes       \\
  RGB-D/T Salient Object Detection (RGB-D/T SOD)               &                                & \yes                          & \yes           &              &             & \yes        &            \\
  RGB-D Semantic Segmentation (RGB-D SS)                    &                                & \yes                          &                &              & \yes        & \yes        &            \\
  \bottomrule[2pt]
\end{tabular}
}
  \setlength{\abovecaptionskip}{0.5em}
  \caption{Attributes of several typical bi-source dense prediction tasks. More details can be found in Sec.~\ref{sec:bisourcetask}.}
  \label{tab:task}
  \vspace{-1em}
\end{table*}

\begin{figure}[!t]
  \centering
  \includegraphics[width=0.7\linewidth]{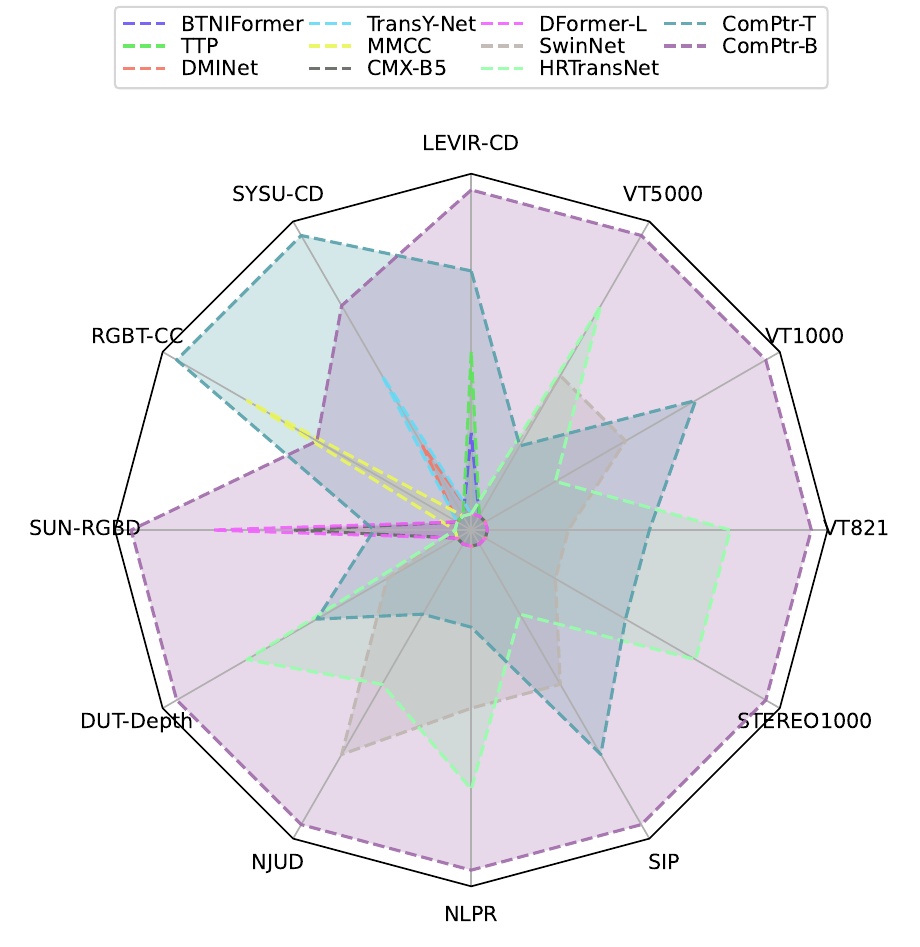}
  \setlength{\abovecaptionskip}{-0.5em}
  \caption{Our method ComPtr achieves the best performance on 12 datasets for 5 bi-source tasks with different properties.}
  \label{fig:all_tasks}
  \vspace{-1em}
\end{figure}

\begin{figure*}[!t]
  \centering
  \includegraphics[width=0.8\linewidth]{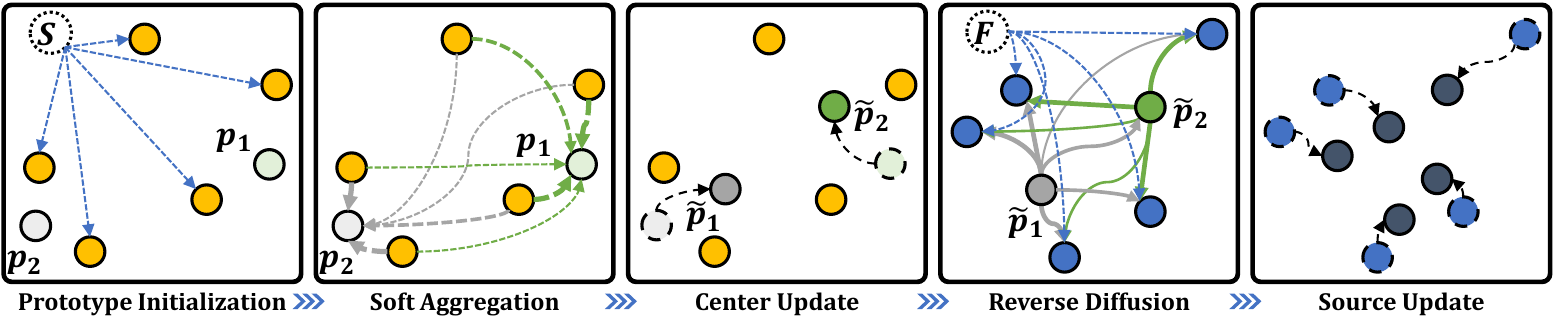}
  \caption{Illustration of our aggregation-diffusion attention (ADA).
  It implements an efficient alternative to the standard attention to build a global interaction between source feature set $S=\{F^1, F^2\}$ and slot features $F$.
  In the feature space, the proxy prototype set $P=\{p_k\}_{k=1}^K$ ($K=2$ in this case) for information interaction is initialized first.
  Based on the similarity, $S$ is adaptively aggregated into these prototypes and they evolve into $\tilde{P}$.
  Next, core semantic concepts, which are compressed into the prototypes, are diffused backward to $F$.
  The thickness of the green and gray arrows represents the relative scale of the weights in the feature reconstruction, where the thicker is larger and vice versa.
  Best viewed in color.
  }
  \label{fig:agg-diff}
\end{figure*}

The global dense interaction plays an important role in bi-source vision tasks.
Due to the explicit long-range correlation mechanism, the transformer-based cross-source modeling scheme {has} become the dominant paradigm~\cite{caver-tip,changedetection-BTNIFormer,rgbtcc-RGBTMMCC,mmseg-TokenFusion,mmseg-MultiMAE,mmseg-DFormer}.
However, its computational complexity increases quadratically with the length of the input token sequence.
This inherent property in the structure of the standard transformer leads to a severe computational burden when dealing with high-resolution features.
To alleviate this problem, we design the aggregation-diffusion attention inspired by the K-means algorithm.
Specifically, several specific proxy prototypes are inserted into the feature interaction path.
It provides an alternative of linear complexity for global information propagation, of which the schematic diagram is shown in Fig.~\ref{fig:agg-diff}.
In the end, the rational design and good collaboration of these proposed components together contribute to the good cross-task compatibility of our model.

  {Our main contributions can be summarized as follows}:
\begin{enumerate}
  \item To the best of our knowledge, it is the first attempt to accomplish the unification of the network architecture for diverse bi-source dense prediction tasks.
  \item We build a simple yet effective complementary transformer, ComPtr, by deconstructing the complementarity property of tasks into two aspects: consistency and difference, which enhances global feature interaction and narrows the gaps among multiple tasks.
  \item A novel proxy prototype bridging strategy is proposed to construct the aggregation-diffusion attention mechanism, which helps to improve and simplify the global information propagation process.
  \item ComPtr achieves state-of-the-art performance on several challenging dataset benchmarks of five representative and distinctive tasks.
\end{enumerate}

\section{Related Work}

\subsection{Bi-source Dense Prediction}
\label{sec:bisourcetask}

The dense prediction field has a large number of branches, and they have a variety of data forms and task objectives.
The bi-source dense prediction tasks can be summarized based on task attributes as listed in Tab.~\ref{tab:task}.
From the forms of inputs, there are two categories: \textit{single-modal data source} from the same type of sensor, and \textit{multi-modal data source} from different types of sensors, which introduces auxiliary information such as depth and infrared radiation.
According to the attributes of objects, they can be divided into three aspects: \textit{class-agnostic}, \textit{single-class}, and \textit{multi-class} tasks.
In addition, their outputs are \textit{complete region segmentation} ({\em i.e.}, confidence or binary map) or \textit{rough point location} ({\em i.e.}, density estimation).
Motivated by these considerations, we select several representative bi-source tasks to explore the general modeling in this work.
As shown in Tab.~\ref{tab:task}, they cover all the aforementioned task attributes.

\subsubsection{Task-specific Methods}

Task-specific methods dominate different tasks, and they use the characteristics and prior information of the task itself as a guide for model design.
In \textit{\textbf{remote sensing change detection}} (RSCD) that aims to explore the dissimilarity of the specific content of bi-temporal images, the exploration of task-inspired difference information has been the main concern of related works.
The pioneering work~\cite{changedetection-FCNN} proposes a fully convolutional siamese neural network to perform change detection and this classical architecture is inherited by many follow-up methods.
And, single-temporal image segmentation~\cite{changedetection-FC-EF-Res,changedetection-DTCDSCN} and edge detection~\cite{changedetection-EGRCNN}, which are closely related to RSCD, are introduced as auxiliaries to construct multi-task learning.
To enhance the mining process of difference information, recent works~\cite{changedetection-IFN,changedetection-BIT,changedetection-ICIF-Net,changedetection-ChangeFormer,changedetection-BTNIFormer,changedetection-TTP,changedetection-DMINet} begin to incorporate the transformer structure and attention mechanism to model contexts within spatial and temporal domains.
In \textit{\textbf{RGB-T crowd counting}}, the thermal modality plays an important role in poor illumination conditions and is further highlighted because of its sensitivity to the human body.
Existing two-stream methods~\cite{rgbtcc-CMCRL-RGBTCC,rgbtcc-DEFNet} and three-stream~\cite{rgbtcc-TAFNet} aggregate complementary information by designing complex cross-modal interaction structures.
Recent work \cite{rgbtcc-MAT} introduces the mutual attention transformer to leverage the cross-modal information.
In \textit{\textbf{RGB-D/T salient object detection}}, the two-stream architecture is the mainstream.
\cite{HDFNet} uses the dynamic convolution module with depth information to guide feature decoding and achieves more flexible and efficient multi-scale cross-modal feature processing.
\cite{ECFFNet-RGBTSOD} designs an effective cross-modal fusion and enables the salient boundary via the bilateral fusion strategy.
Recently, by introducing the transformer-based dense interaction component, \cite{TriTransNet-RGBDSOD,rgbdsod-VST,rgbdtsod-SwinNet,caver-tip,rgbdtsod-HRTransNet} achieve superior results.
In \textit{\textbf{RGB-D semantic segmentation}}, the geometric prior of depth information is fully depicted and utilized.
\cite{mmseg-3DGNN} builds a k-nearest neighbor graph based on position and depth information.
\cite{mmseg-SGNet} {uses} the depth modality to help the convolution layer adjust the receptive field and adapt to geometric transformations.
\cite{mmseg-ShapeConv} designs a depth-specific convolution layer and pays more attention to shape information by re-weighting the shape component and the base component of the depth feature.
Besides, in~\cite{mmseg-SA-Gate,mmseg-PGDENet,mmseg-Asymformer}, the gating/attention mechanism is introduced to selectively aggregate the informative features.
All the above-mentioned schemes, without exception, focus on task-specific information mining.
They make full use of task-prior knowledge to obtain more inductive bias.

\subsubsection{Task-generic Methods}

The task-generic design can improve the efficiency of data utilization and take a solid step towards a more intelligent algorithm by integrating knowledge of various tasks.
The existing task-generic models can be roughly classified into three categories: the large-scale pre-training task-independent architecture~\cite{BERT,GPT-3,mmseg-MultiMAE}, the multi-task joint learning framework~\cite{generalmodel-Unicorn} and the transferable general single-task architecture~\cite{rgbdtsod-SwinNet,caver-tip,mmseg-CMX,mmseg-TokenFusion}.
This paper belongs to the last one.
  {Although several recent approaches~\cite{rgbdtsod-SwinNet,caver-tip,mmseg-CMX,mmseg-TokenFusion} have attempted to explore the cross-task unified architecture, they mainly focus on unifying homogeneous bi-source tasks ({\em e.g.}, cross-modal salient object detection~\cite{rgbdtsod-SwinNet,caver-tip,rgbdtsod-HRTransNet} and semantic segmentation~\cite{mmseg-DFormer,mmseg-CMX,mmseg-CMNeXt,mmseg-TokenFusion}), but lack consideration for unifying heterogeneous bi-source tasks ({\em e.g.}, tasks with different attributes as shown in Tab.~\ref{tab:task}).
    \textbf{There is a large room in the extension of task concepts, data forms, and object attributes, which is exactly what we focus on in this work.}}

\subsection{Bi-source Transformer-based Methods}

Long-range context information is crucial for accurate scene understanding, which has been verified in many existing works~\cite{DilatedConvolution,Deeplab,LargeKernel}.
Due to the built-in long-range dependence modeling,%
the non-local operation like the non-local block~\cite{NonLocalNet} and self-attention in transformer~\cite{ViT,ViT-Adapter,Swin}, has become an important component for capturing the global context cues.
Recently, several transformer-based approaches have been proposed for bi-source dense prediction tasks.
Most of those methods~\cite{changedetection-DMINet,rgbdsod-VST,rgbdtsod-SwinNet,caver-tip,rgbtcc-RGBTMMCC,rgbtcc-MAT} follow the design pattern of using independent encoders for different sources.
They apply vision transformers to construct hierarchical representations and ignore the complementarity characterization in the encoder stage.
Moreover, their interaction units based on standard attention also limit the scalability of their structure~\cite{changedetection-DMINet,rgbdsod-VST,rgbtcc-MAT,caver-tip,mmseg-MultiMAE,mmseg-TokenFusion}.
With less inductive bias and relying only on data-driven learning, these schemes may struggle to effectively perceive key cues in complex contexts~\cite{ViT}.
Different from them, we use the proxy prototype mechanism inspired by the clustering algorithm to transform the global interaction of complementary information into two sequential steps of aggregation and diffusion.
This design effectively improves the scalability of the proposed architecture for different tasks and plays a positive role in realizing a general model for diverse bi-source dense prediction tasks.

\subsection{{Token Reduction for Transformer}}

{%
  The transformer suffers from quadratic computational complexity with respect to the input sequence length, especially when handling multi-scale and high-resolution features.
  While recent studies \cite{Wu_2023_ICCV,Dou_2023_ICCV,Grainger_2023_CVPR} mitigate this via static non-linear convolutional layers to compress redundant image tokens, their inherent spatial structure disruption and static assignment mechanisms limit applicability to diverse bi-source dense prediction tasks that we focus on.
  In contrast, our K-means-inspired approach introduces a similarity-driven dynamic assignment strategy that adaptively preserves the spatial structure of image features while perceiving global content correlations.
}

\begin{figure*}[!t]
  \centering
  \includegraphics[width=0.8\linewidth]{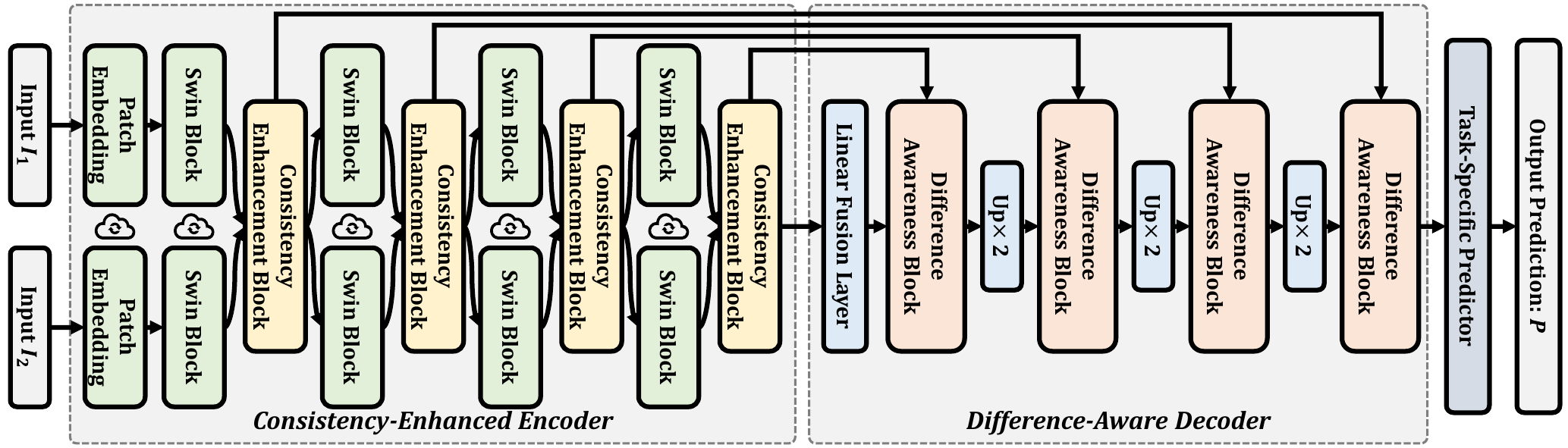}
  \caption{Overview of the proposed ComPtr framework.
    It follows the two-branch encoder-decoder architecture which is commonly used in diverse bi-source dense prediction tasks.
    \includegraphics[height=1em]{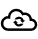} indicates that the corresponding module shares parameters.
    The prediction $P$ of the model with the inputs $I_1$ and $I_2$ will be supervised by the ground truth $G$.
  }
  \label{fig:net}
\end{figure*}

\section{Methodology}

In this section, we first present the overall structure of the proposed ComPtr and then introduce the details of the different components in sequence.

\subsection{Overall Architecture}\label{sec:overallarchitecture}

Following the two-branch encoder-decoder architecture which is commonly used in diverse bi-source dense prediction tasks~\cite{changedetection-FCNN,TriTransNet-RGBDSOD,caver-tip,ECFFNet-RGBTSOD,VT5000-ADF}, we design a \textbf{comp}lementary \textbf{tr}ansformer framework, \textbf{ComPtr}.
Considering that complementarity plays an important role in the bi-source dense prediction task, this concept is decomposed into two parts, {\em i.e.}, consistency and difference, in the proposed model.
And they lead the improvement and design of the encoder and decoder respectively.
As shown in Fig.~\ref{fig:net}, ComPtr is composed of these basic components: a \textit{parameter-shared feature extractor}, several \textit{consistency enhancement and difference awareness blocks}, and a \textit{light task-specific predictor}.

\subsubsection{Consistency-Enhanced Encoder}

We directly choose the relatively frequently-used Swin Transformer~\cite{Swin} to realize the function of feature extraction which consists of several \textit{non-overlap patch embedding layers} and \textit{swin transformer blocks} built on the shifted window based self-attention.
\textbf{The siamese design with shared parameters is also introduced here} like~\cite{changedetection-FCNN} to simplify the model design and reduce the number of parameters.
Besides, the proposed consistency enhancement component is inserted into the basic feature extractor to build the consistency enhanced encoder as depicted on the left of Fig.~\ref{fig:net}.
{It allows each stream to be strengthened with these important consistency cues, while preserving information in the original feature stream due to the residual transformer form with identity links.}

\subsubsection{Difference-Aware Decoder}

In the decoder, the deepest two-stream features are merged by a simple linear fusion layer only containing normalization and linear layers.
The result is delivered to a cascaded difference awareness block and intersects with the shallow encoder features, which absorbs the difference information at the specific scale between the two streams.
Together with the subsequent stacked interpolation operations, difference awareness blocks, and task-specific predictor, they form the difference-aware decoder as shown on the right of Fig.~\ref{fig:net}.
{On the basis of features enhanced by consistency-related patterns, the model continuously absorbs inter-source difference-specific representations of features from different scales of the encoder and integrates them into the feature decoding stream internally, which is then used for the final task-specific object decoding.}

\subsubsection{Light Task-specific predictor}

To adapt to different types of tasks, we introduce several lightweight task-specific predictors by simply combining interpolation, normalization, activation, and linear layers.

\begin{figure*}[!t]
  \centering
  \includegraphics[width=0.8\linewidth]{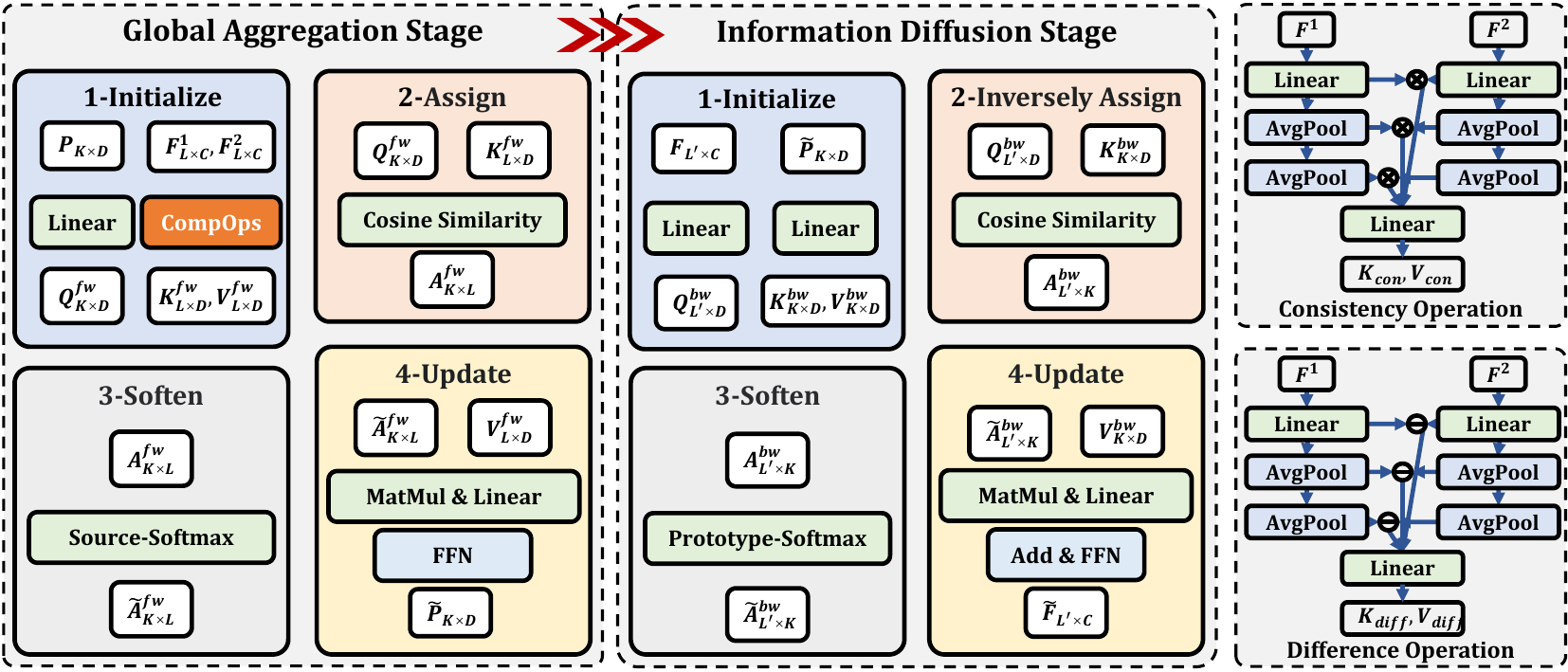}
  \caption{Details of the proposed aggregation-diffusion attention mechanism (ADA) which {constructs} the interaction between the source features $F^1, F^2 \in \mathbb{R}^{L \times C}$ and the slot feature $F \in \mathbb{R}^{L' \times C}$ mediated by the learnable proxy prototype $P \in \mathbb{R}^{K \times D}$.
    It contains two stages executed sequentially: a forward global aggregation stage (``fw'') and a backward information diffusion stage (``bw'').
    ``CompOps'' represents the complementarity operations as shown on the right: the multi-granularity consistency operation and the hierarchical difference operation.
    \includegraphics[width=1.5ex]{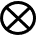} and \includegraphics[width=1.5ex]{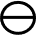} generate the element-wise product and the absolute difference of the two inputs.
    Some normalization and activation layers are dropped here for notational brevity.
  }
  \label{fig:attention}
\end{figure*}

\subsection{Complementary Transformer}\label{sec:complementarytransoformer}

The consistency enhancement and difference awareness blocks are all based on a novel and efficient global aggregation-diffusion attention mechanism (ADA), but there are obvious differences in the details.

\subsubsection{Aggregation-Diffusion Attention}\label{sec:aggdffattn}

\begin{figure}[t]
  \begin{minipage}[t]{\linewidth}
    \begin{algorithm}[H]
      \footnotesize
      \caption{Standard K-means Algorithm}
      \label{alg:kmeans}
      \begin{algorithmic}[1]
        \State Initialize cluster centroids $\{\mu_k\}_{k=1}^{K}$ for $\{x_m\}_{m=1}^{M}$
        \Repeat
        \For{$m=1, \dotsc, M$}{\Comment \texttt{Assign}}
        \State $c_m$ = index of cluster centroid closest to $x_m$
        \EndFor
        \For{$k=1, \dotsc, K$}{\Comment \texttt{Update}}
        \State $\mu_k$ = average of points assigned to cluster $k$
        \EndFor
        \Until{converged}
      \end{algorithmic}
    \end{algorithm}
  \end{minipage}
  \par
  \begin{minipage}[t]{\linewidth}
    \begin{algorithm}[H]
      \footnotesize
      \caption{Aggregation-Diffusion Attention}
      \label{alg:ada}
      \begin{algorithmic}[1]
        \State Initialize learnable proxy prototypes $\{p_k\}_{k=1}^{K}$ for source features $\{f_i\}_{i=1}^{L}$ and slot features $\{f'_i\}_{i=1}^{L'}$
        \Repeat
        \For{$i=1, \dotsc, L$}{\Comment \texttt{Assign}}
        \State $a^{fw}_i$ = similarity between $f_i$ and $\{p_k\}_{k=1}^{K}$
        \EndFor
        \For{$k=1, \dotsc, K$}{\Comment \texttt{Update}}
        \State $\tilde{p}_k$ = weighted average of $\{f_i\}_{i=1}^{L}$ assigned to $p_k$
        \EndFor
        \For{$k=1, \dotsc, K$}{\Comment \texttt{Inversely Assign}}
        \State $a^{bw}_k$ = similarity between $\tilde{p}_k$ and $\{f'_i\}_{i=1}^{L'}$
        \EndFor
        \For{$i=1, \dotsc, L'$}{\Comment \texttt{Inversely Update}}
        \State $f'_i$ = weighted average of $\{\tilde{p}_k\}_{k=1}^{K}$ assigned to $f'_i$
        \EndFor
        \Until{converged}
      \end{algorithmic}
    \end{algorithm}
  \end{minipage}
\end{figure}

In order to build a more effective dense interaction between the features from the two sources to promote the mining and dissemination of global complementary information, we introduce the transformer architecture.
Although it has powerful information modeling and interaction capabilities, its global dense interaction form also causes a non-negligible quadratic computational complexity with respect to the length of the input sequence.
Inspired by the K-means clustering algorithm, we introduce the proxy prototype as the mediator for global information propagation, and the computational logic for both is shown in Alg.~\ref{alg:kmeans} and Alg.~\ref{alg:ada}, respectively.
And the straight-through dense interaction of the standard attention is decoupled and reconstructed into two different stages: the forward global aggregation and the backward information diffusion as depicted in Fig.~\ref{fig:attention}.

In the \textbf{aggregation stage}, the attribute-specific key and value $K^{fw}, V^{fw} \in \mathbb{R}^{L \times D} (L = H \times W)$ are generated from two-stream source feature set $S=\{F^i\}^2_{i=1}, F^i \in \mathbb{R}^{L \times C}$ by the complementarity operation $\mathtt{CompOps}$.
And $K$ learnable $D$-dimensional proxy prototype vectors $P = \{p_i\}_{i=1}^{K}$ are also transformed to the same channel dimension $D$ by a linear projection weight $W^{fw}_{Q}$.
The above process can be formulated as:
\begin{gather}
  K^{fw}, V^{fw} = \mathtt{CompOps}(F^1, F^2), \label{equ:compops} \\
  Q^{fw} = P W^{fw}_{Q}, \quad W^{fw}_{Q} \in \mathbb{R}^{D \times D}.
\end{gather}
Inspired by the similarity-based information aggregation in K-means, the pair-wise $\mathtt{cosine}$ similarity between the source embedding and the proxy prototype $P$ {serves} as the basis for assigning the global information in the aggregation process, which can be expressed mathematically as:
\begin{equation}
  \begin{gathered}
    A^{fw} = \mathtt{cosine}(Q^{fw}, K^{fw}).
  \end{gathered}
\end{equation}
The next steps are similar to the centralization process in K-means.
Different from the hard sample selection mechanism ($\mathtt{argmax}$) in K-means, $\mathtt{softmax}$ is chosen to soften the process and ensure the differentiability of the calculation.
So we can get the normalized similarity as follows:
\begin{equation}
  \begin{gathered}
    \tilde{A}^{fw} = \mathtt{softmax}(A^{fw}).
  \end{gathered}
\end{equation}
And a similarity-based transformation is introduced to integrate the more important cues from the transformed input $V^{fw}$ into the proxy prototype $P$.
The updated proxy prototype $\tilde{P} \in \mathbb{R}^{K \times D}$ can be obtained as follows:
\begin{equation}
  \begin{gathered}
    \tilde{P} = \mathtt{FFN}((\tilde{A}^{fw} V^{fw}) W^{fw}_{O}), \quad W^{fw}_{O} \in \mathbb{R}^{D \times D},
  \end{gathered}
\end{equation}
where $\mathtt{FFN}$ is a forward feed network containing normalization, linear and activation layers.

And in the \textbf{diffusion stage}, the proxy prototype $\tilde{P}$, acting as a mediator, {diffuses} these global semantics back into the original image space, and this procedure can also be considered as reversed K-means.
At this time, the image feature $F \in \mathbb{R}^{L' \times C}$, {\em i.e.}, the diffusion slot, can supplement the complement information from the prototype $\tilde{P}$ based on the similarity relationship.
After the specific linear transformation by $W^{bw}_{Q}$ and $W^{bw}_{K}$, both $F$ and $\tilde{P}$ are involved in the calculation of the correlation, which can generate the similarity $A^{bw}$ after the $\mathtt{cosine}$ operation as following:
\begin{equation}
  \begin{gathered}
    Q^{bw} = F W^{bw}_{Q}, \quad W^{bw}_{Q} \in \mathbb{R}^{C \times D}, \\
    K^{bw} = \tilde{P} W^{bw}_{K}, \quad W^{bw}_{K} \in \mathbb{R}^{D \times D}, \\
    A^{bw} = \mathtt{cosine}(Q^{bw}, K^{bw}).
  \end{gathered}
\end{equation}
The $\mathtt{softmax}$ function is imposed on $A^{bw}$ to normalize the importance of different prototypes and modulate the subsequent feature reconstruction based on $\tilde{P}$.
And it can be mathematically expressed as follows:
\begin{equation}
  \begin{gathered}
    \tilde{A}^{bw} = \mathtt{softmax}(A^{bw}), \\
    V^{bw} = \tilde{P} W^{bw}_{V}, \quad W^{bw}_{V} \in \mathbb{R}^{D \times D}, \\
    Z^{bw} = \tilde{A}^{bw} V^{bw} W^{bw}_{O}, \quad W^{bw}_{O} \in \mathbb{R}^{D \times C}.
  \end{gathered}
\end{equation}
And the residual connection and learnable vector $\gamma \in \mathbb{R}^{D}$ are introduced to balance the attention output $Z^{bw}$ and the input slot feature $F$, where $\gamma$ is initialized with the zero value to ensure the original feature distribution in the early stage of training as~\cite{ViT-Adapter}:
\begin{equation}
  \begin{gathered}
    \tilde{F} = F + \mathtt{FFN}(F + \gamma \odot Z^{bw}),
  \end{gathered}
\end{equation}
where $\odot$ denotes the element-wise multiplication.
The successive aggregation and diffusion phases realize the global interaction based on the proxy prototype.

\begin{figure*}[t]
  \centering
  \includegraphics[width=0.8\linewidth]{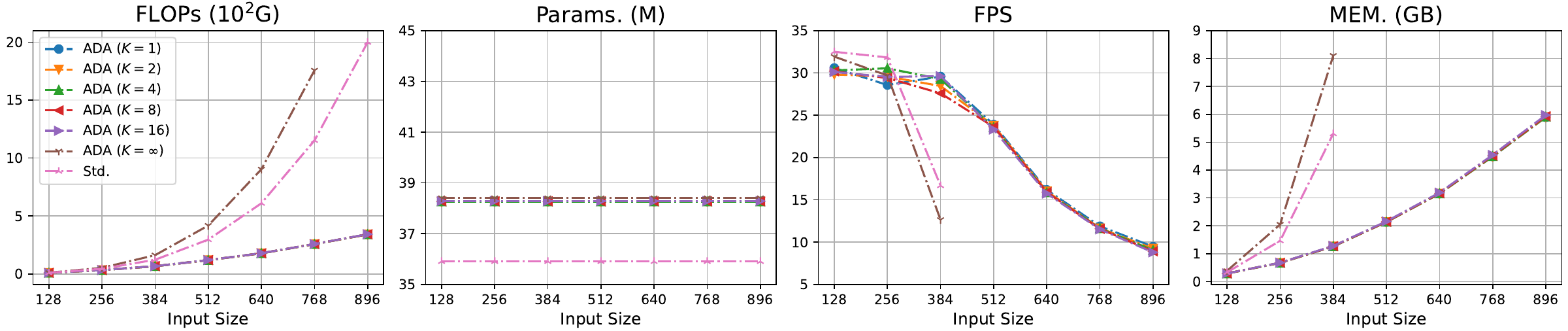}
  \caption{Comparison of the proposed ADA with the standard attention form ``Std.'' in terms of the amount of computation (``{FLOPs}''),  the number of parameters (``Params.''), the inference speed (``FPS''), and the peak memory usage (``MEM.'').
  And these results are collected in the \textit{inference mode} with an NVIDIA RTX 2080 Ti GPU, PyTorch 1.12, and Ubuntu 18.04, focusing on the inference efficiency of the model.
  It is worth noting that the FPS and MEM. of ``ADA (K=$\infty$)'' and ``Std.'' cannot be evaluated with input sizes larger than $384 \times 384$ due to their huge memory requirements.
  More details can be found in Sec.~\ref{sec:compwithstdattn} and Sec.~\ref{sec:num_k}.
  }
  \label{fig:efficiency}
\end{figure*}

\subsubsection{{Comparison with Standard Attention}}
\label{sec:compwithstdattn}

For a $D$-dimensional image token sequence of length $L = H \times W$, the computational complexity and the memory footprint of the standard attention are approximately $O(L^{2} \times D)$ and $O(L^{2})$.
Compared with the standard attention, the proposed proxy prototype bridging strategy effectively reduces the computation and memory of dense global interaction to linear complexity ($O(L \times K \times D + L \times K \times D)$ and $O(L \times K + L \times K)$), where $K$ is the number of proxy prototypes.
Considering that $K$ is a small constant, {\em i.e.} $K \ll L$, the proposed aggregation-diffusion attention is more efficient than the standard form.
In order to visualize the difference in inference efficiency, we show the comparison of the proposed ADA with the standard attention for different resolutions of the input in Fig.~\ref{fig:efficiency}, and more details and analyses can be found in Sec.~\ref{sec:num_k}.

\subsubsection{Consistency Enhancement Block (CEB)}
\label{sec:consistencyblock}

By introducing the consistency operation into the proposed aggregation-diffusion attention, we design the CEB to strengthen the common visual cues and help the model obtain a more powerful data representation.
  {The CEB in the encoder focuses on consistency enhancement to strengthen common object representations (\textit{e.g.}, general scene understanding), which aligns with and enhances the encoder's role in extracting high-level semantics.}
And the consistency-specialized key and value embeddings $K_{con}$ and $V_{con}$ can be obtained by the multi-granularity element-wise multiplication of the two image source feature inputs $F^1$ and $F^2$ from different streams as shown on the top right-hand corner of Fig.~\ref{fig:attention}.
They are used as $K^{fw}$ and $V^{fw}$ in Equ.~\ref{equ:compops}.
Specifically, the consistency operation is composed of several normalization, linear, and pooling layers.
With the aid of two parameter-free average pooling layers, we can obtain products $\{F^i_{con}\}_{i=1}^3$ from three different scales.
After {being concatenated} and transformed along the channel, a complete multi-scale consistency representation is generated from the inputs.
It is split into $K_{con}$ and $V_{con}$ along the channel which participate in the update process of the proxy prototype $P$.
In the diffusion stage, the slot feature $F$ with $L' = 2HW$ is the concatenation of the image features $F^1$ and $F^2$ along the {flattened} spatial dimension.
It is reconstructed independently to $\tilde{F}$ based on the updated proxy prototype $\tilde{P}$.
And these enhanced features $\tilde{F}^1$ and $\tilde{F}^2$ stored in the slot $\tilde{F}$ replace original $F^1$ and $F^2$ as the input of the next swin transformer block.

\subsubsection{Difference Awareness Block (DAB)}
\label{sec:differenceblock}

{The DAB embedded in the decoder emphasizes difference awareness to recover task-related details (\textit{e.g.}, cross-modal structural compensation in RGB-D SOD and temporal variations in RSCD), which matches the decoder's function of refining spatial details.}
This block differs from the CEB in two aspects, one is the generation of embeddings $K_{diff}$ and $V_{diff}$ based on the source features, and the other is the global diffusion way of the assigned information.
As shown on the bottom right-hand corner of Fig.~\ref{fig:attention}, a hierarchical difference operation containing normalization, linear and pooling layers are embedded here to obtain the multi-granularity absolute difference representation $K_{diff}$ and $V_{diff}$, which are used as $K^{fw}$ and $V^{fw}$ in Equ.~\ref{equ:compops}.
Besides, the slot feature $F$ with $L' = HW$ in the diffusion stage is a compact hybrid representation, which comes from the initial fusion of the encoder features $F^1$ and $F^2$, and the deeper DAB feature $\tilde{F}'$ through several stacked linear layers.
The output slot feature $\tilde{F}$ of the block is fed into the subsequent module and involved in the decoding process of the high-resolution prediction.

\subsubsection{{Expectation-Based Operation Analysis}}

{Expectation analysis provides a concise yet profound framework for understanding the intrinsic behavior of mathematical operations in signal processing and statistical modeling.
So we leverage expectation as the core tool to better analyze both operations.
We assume that all signals follow Gaussian distributions, which is a common and reasonable assumption in signal processing and statistics due to the \textit{Central Limit Theorem}.
Let $F_1 = C + D_1$ and $F_2 = C + D_2$ be the two input signals from different sources.
$C \sim \mathcal{N}(\mu_C, \sigma_C^2)$ represents the consistent feature, which is common to both signals.
$D_1 \sim \mathcal{N}(0, \sigma_{D_1}^2)$ and $D_2 \sim \mathcal{N}(0, \sigma_{D_2}^2)$ represent the difference features specific to each signal.
Here, for simplicity, we assume that $D_1$ and $D_2$ are zero-mean and that $C$, $D_1$ and $D_2$ are independent of each other.
For \textbf{the element-wise product} of two signals $F_1$ and $F_2$, the expectation is:
$E[F_1 F_2] = E[(C+D_1) (C+D_2)] = E[C^2] + E[C]E[D_1] + E[C]E[D_2] + E[D_1]E[D_2] = E[C^2]$.
Since $C$ is the dominant consistent feature and cross-terms vanish under the independence assumption, the dominant term is $E[C^2]$, indicating that the element-wise product emphasizes consistency.
For \textbf{the absolute difference} of two signals $F_1$ and $F_2$, the expectation is:
$E[|F_1 - F_2|] = E[| (C+D_1) - (C+D_2) |] = E[|D_1 - D_2|]$.
The focus of this operation is on the difference between $D_1$ and $D_2$, effectively capturing difference information.
In conclusion, the element-wise product operation emphasizes the consistent feature $C$ while the absolute difference operation focuses on the difference information represented by $D_1$ and $D_2$.}

\subsubsection{Loss Function}

For a fair comparison with existing approaches, we use the commonly used loss functions for each task.
In remote sensing change detection, RGB-D/T salient object detection, and RGB-D semantic segmentation tasks, the binary or multi-class cross-entropy loss is used to supervise the training of the model.
In RGB-T crowd counting, we follow the loss form of~\cite{rgbtcc-RGBTMMCC} containing the distribution matching loss proposed in~\cite{cc-dm} for density map estimation and an L1 loss for counting regression.

\section{Experiments}

To verify the generality, transferability and effectiveness of the proposed architecture, we select four types of important bi-source tasks with large diversity:
remote sensing change detection (RSCD),
RGB-T crowd counting (RGB-T CC),
RGB-D/T salient object detection (RGB-D/T SOD),
and RGB-D semantic segmentation (RGB-D SS).
Specific experimental details are presented in the following sections.

\subsection{Basic Training Settings}

In all experiments, we use Swin Transformer~\cite{Swin} pre-trained from ImageNet~\cite{ImageNet} as the initial parameters of the feature extractor for all tasks, while other parts are randomly initialized by PyTorch.
The three-channel image is normalized, while the single-channel one only replicates three times along the channel after dividing by 255.
The AdamW~\cite{AdamW} optimizer is consistently used due to its suitability for the transformer architecture.
The settings of other hyper-parameters for different tasks mostly follow those of the existing algorithms~\cite{changedetection-DMINet,rgbtcc-RGBTMMCC,caver-tip,rgbdtsod-SwinNet,SPNet-RGBDSOD-journal,ECFFNet-RGBTSOD,mmseg-SA-Gate} to achieve a more fair comparison, and more task-specific details are shown in the corresponding sections.

\subsection{Remote Sensing Change Detection (RSCD)}\label{sec:rscd-setting}

\noindent\textbf{Datasets.}
Two commonly used large-scale data benchmarks are used to evaluate performance, namely LEVIR-CD~\cite{changedetection-LEVIR} and SYSU-CD~\cite{changedetection-SYSU}.
\textbf{LEVIR-CD} is a large-scale and very-high-resolution (VHR) RSCD dataset collected via Google Earth API.
It contains 637 pairs of real bi-temporal RGB image patches with a time span of 5-14 years, a spatial resolution of 0.5 m/pixel and a total of 31333 individual change buildings.
The dataset is split into training/validation/testing sets with 7120/1024/2048 samples.
The recently proposed \textbf{SYSU-CD} is another important and challenging benchmark containing 20000 pairs of 0.5 m/pixel aerial images taken between the years 2007 and 2014 in Hong Kong.
These images are divided into training/validation/testing sets with 12000/4000/4000 image pairs.
Following the existing methods~\cite{changedetection-DMINet,changedetection-ICIF-Net,changedetection-BIT}, the model is trained, validated, and tested independently on each dataset.
And bi-temporal image pairs with a shape of $256 \times 256$ and a batch size of 16 are augmented using random scaling, affine transformation, flipping, and color jitter during training.
For the experiments on the LEVIR-CD dataset, we train the model for 200 epochs and use the cosine scheduler with an initial learning rate of 0.0002 and a weight decay of 0.0005.
On SYSU-CD, to alleviate the over-fitting problem, we introduced the one-cycle scheduler~\cite{onecycle} with a large learning rate of 0.001 and a weight decay of 0.0005, which can play a role of regularization, and the number of training epochs is reduced to a quarter of the one on LEVIR-CD.

\noindent\textbf{Metrics.}
For the comparison between the change prediction map and the ground truth, we introduce five common binary image similarity metrics, including
\textit{precision} ($\mathrm{Pre.} = \frac{\text{TP}}{\text{TP} + \text{FP}}$),
\textit{recall} ($\mathrm{Rec.} = \frac{\text{TP}}{\text{TP} + \text{FN}}$),
\textit{F1-score} ($\mathrm{F1}   = \frac{2\text{TP}}{2\text{TP} + \text{FN} + \text{FP}}$),
\textit{intersection over union} ($\mathrm{IoU}  = \frac{\text{TP}}{\text{TP} + \text{FP} + \text{FN}}$) and
\textit{overall accuracy} ($\mathrm{OA}   = \frac{\text{TP} + \text{TN}}{\text{TP} + \text{TN} + \text{FP} + \text{FN}}$).
TP, TN, FP, and FN are the number of true positive, true negative, false positive, and false negative samples in the binary prediction, respectively.
Since F1 and IoU implicitly consider both Pre. and Rec., they can reflect the performance of the model more generally.

\begin{figure}[!t]
  \begin{minipage}{\linewidth}
    \centering
    \resizebox{\linewidth}{!}{%
      \rowcolors{2}{gray!10}{white}
      \begin{threeparttable}
  \begin{tabular}{l|rr|l|*5{c}}
    \toprule[2pt]
    Methods                                                 & {Params.}         & {FPS}           & Backbone                                         & Pre.~$\uparrow$ & Rec.~$\uparrow$ & F1~$\uparrow$  & IoU~$\uparrow$ & OA~$\uparrow$  \\
    \midrule[1pt]
    FC-EF$_{18}$~\cite{changedetection-FCNN}                & {1.4M}            & {400.2}         & UNet~\cite{UNet}                                 & 85.56           & 78.76           & 82.02          & 69.52          & 98.24          \\
    FC-Siam-Di$_{18}$~\cite{changedetection-FCNN}           & {1.4M}            & {284.8}         & UNet~\cite{UNet}                                 & 91.53           & 78.24           & 84.36          & 72.95          & 98.52          \\
    FC-Siam-Conc$_{18}$~\cite{changedetection-FCNN}         & {1.5M}            & {293.0}         & UNet~\cite{UNet}                                 & 84.17           & 79.49           & 81.77          & 69.16          & 98.19          \\
    FresUNet$_{19}$~\cite{changedetection-FC-EF-Res}        & {1.1M}            & {235.0}         & UNet~\cite{UNet}                                 & 89.86           & 86.41           & 89.05          & 80.26          & 98.92          \\
    IFNet$_{20}$~\cite{changedetection-IFN}                 & {50.7M}           & {68.4}          & VGG-16~\cite{VGG}                                & \best{94.39}    & 82.42           & 88.01          & 78.58          & 98.86          \\
    DTCDSCN$_{21}$~\cite{changedetection-DTCDSCN}           & {41.1M}           & {71.0}          & SE-ResNet-32~\cite{SENet}                        & 89.25           & 85.68           & 87.43          & 77.67          & 98.75          \\
    SNUNet$_{22}$~\cite{changedetection-SNUNet}             & {12.0M}           & {62.0}          & NestedUNet~\cite{UNet++}                         & 89.18           & 87.17           & 88.16          & 78.83          & 98.82          \\
    BIT$_{22}$~\cite{changedetection-BIT}                   & {11.9M}           & {86.1}          & ResNet-18~\cite{Resnet}                          & 89.24           & 89.37           & 89.31          & 80.68          & 98.92          \\
    EGRCNN$_{22}$~\cite{changedetection-EGRCNN}             & {15.7M}           & {61.1}          & UNet~\cite{UNet}                                 & 92.83           & 86.46           & 89.59          & 81.12          & 98.92          \\
    MSPSNet$_{22}$~\cite{changedetection-MSPSNet}           & {2.2M}            & {132.7}         & MSSNN~\cite{changedetection-MSPSNet}             & 91.38           & 87.08           & 89.18          & 81.36          & {99.03}        \\
    ICIFNet$_{22}$~\cite{changedetection-ICIF-Net}          & {25.8M}           & {27.0}          & ResNet-18~\cite{Resnet}+PVT-V2-B1~\cite{PVTv2}   & 91.32           & 88.64           & 89.96          & 81.75          & 98.99          \\
    ChangeFormer$_{22}$~\cite{changedetection-ChangeFormer} & {41.0M}           & {33.1}          & ChangeFormer~\cite{changedetection-ChangeFormer} & 92.05           & 88.80           & 90.40          & 82.48          & 99.04          \\
    SACNet$_{22}$~\cite{changedetection-SACNet}             & {N/A\tnote{1}}    & {N/A\tnote{1}}  & VGG-16~\cite{VGG}                                & 91.66           & 89.82           & 90.73          & 83.03          & 98.92          \\
    FTN$_{22}$~\cite{changedetection-FTN}                   & {168.5M}          & {24.5}          & Swin-B~\cite{Swin}                               & 92.71           & 89.37           & 91.01          & 83.51          & 99.06          \\
    TransY-Net$_{23}$~\cite{changedetection-TransYNet}      & {168.5M\tnote{2}} & {24.5\tnote{2}} & Swin-B~\cite{Swin}                               & 92.90           & 89.35           & 91.90          & 83.64          & 99.07          \\
    BTNIFormer$_{23}$~\cite{changedetection-BTNIFormer}     & {N/A\tnote{1}}    & {N/A\tnote{1}}  & BTNIFormer~\cite{changedetection-BTNIFormer}     & 93.32           & 90.77           & 92.02          & 85.23          & \third{99.20}  \\
    TTP$_{23}$~\cite{changedetection-TTP}                   & {348.7M}          & {0.8}           & SAM (ViT-L/16)~\cite{mmseg-SAM}                  & 93.00           & \best{91.70}    & \third{92.10}  & \third{85.60}  & \third{99.20}  \\
    DMINet$_{23}$~\cite{changedetection-DMINet}             & {6.8M}            & {141.1}         & ResNet-18~\cite{Resnet}                          & 92.52           & 89.95           & 90.71          & 82.99          & 99.07          \\
    \midrule[0.5pt]
    SwinNet$^{\daleth}_{22}$~\cite{rgbdtsod-SwinNet}        & {199.2M}          & {26.0}          & Swin-B~\cite{Swin}                               & 92.52           & 89.28           & 90.87          & 83.27          & 99.09          \\
    DFormer-L$^{\daleth}_{24}$~\cite{mmseg-DFormer}         & {39.0M}           & {35.5}          & DFormer-L~\cite{mmseg-DFormer}                   & 92.35           & 89.17           & 90.73          & 83.04          & 99.07          \\
    \midrule[0.5pt]
    ComPtr-T                                                & {38.3M}           & {30.8}          & Swin-T~\cite{Swin}                               & \third{94.29}   & \third{91.00}   & \second{92.62} & \second{86.25} & \second{99.26} \\
    ComPtr-B                                                & {105.7M}          & {21.6}          & Swin-B~\cite{Swin}                               & \second{94.30}  & \second{91.15}  & \best{92.70}   & \best{86.39}   & \best{99.27}   \\
    \bottomrule[2pt]
  \end{tabular}
  \begin{tablenotes}
    \item[1] The code is not publicly available, and the original paper does not provide complete information.
    \item[2] The released code of TransY-Net~\cite{changedetection-TransYNet} is the same with its conference version FTN~\cite{changedetection-FTN}.
  \end{tablenotes}
\end{threeparttable}

    }
    \captionof{table}{Comparison on LEVIR-CD~\cite{changedetection-LEVIR} for RSCD.
      ``$\daleth$'' denotes the methods retrained by us as mentioned in Sec.~\ref{sec:sota-cmp}.
      \textbf{Colors \textcolor{reda}{red}, \textcolor{mygreen}{green} and \textcolor{myblue}{blue} in all tables represent the first, second and third ranked results.}
    }
    \label{tab:rscd-levir}
  \end{minipage}
  \par
  \begin{minipage}{\linewidth}
    \vspace{1em}
    \centering
    \resizebox{\linewidth}{!}{%
      \rowcolors{2}{gray!10}{white}
      \begin{tabular}{l|rr|l|*5{c}}
  \toprule[2pt]
  Methods                                            & {Params.} & {FPS}   & Backbone                                       & Pre.~$\uparrow$ & Rec.~$\uparrow$ & F1~$\uparrow$  & IoU~$\uparrow$ & OA~$\uparrow$  \\
  \midrule[1pt]
  FC-EF$_{18}$~\cite{changedetection-FCNN}           & {1.4M}    & {400.2} & UNet~\cite{UNet}                               & 72.28           & 72.85           & 72.57          & 56.94          & 87.01          \\
  FC-Siam-Di$_{18}$~\cite{changedetection-FCNN}      & {1.4M}    & {284.8} & UNet~\cite{UNet}                               & 84.94           & 51.45           & 64.08          & 47.15          & 86.40          \\
  FC-Siam-Conc$_{18}$~\cite{changedetection-FCNN}    & {1.5M}    & {293.0} & UNet~\cite{UNet}                               & 83.03           & 71.62           & 76.89          & 62.45          & 89.35          \\
  FresUNet$_{19}$~\cite{changedetection-FC-EF-Res}   & {1.1M}    & {235.0} & UNet~\cite{UNet}                               & 78.56           & 74.96           & 76.38          & 61.85          & 89.12          \\
  IFNet$_{20}$~\cite{changedetection-IFN}            & {50.7M}   & {68.4}  & VGG-16~\cite{VGG}                              & 79.59           & 73.58           & 76.53          & 61.91          & 89.17          \\
  DTCDSCN$_{21}$~\cite{changedetection-DTCDSCN}      & {41.1M}   & {71.0}  & SE-ResNet-32~\cite{SENet}                      & 83.19           & 77.25           & 80.11          & 66.82          & 90.96          \\
  SNUNet$_{22}$~\cite{changedetection-SNUNet}        & {12.0M}   & {62.0}  & NestedUNet~\cite{UNet++}                       & 83.49           & 76.37           & 79.77          & 66.35          & 90.87          \\
  BIT$_{22}$~\cite{changedetection-BIT}              & {11.9M}   & {86.1}  & ResNet-18~\cite{Resnet}                        & 78.73           & 75.68           & 77.17          & 62.83          & 89.44          \\
  EGRCNN$_{22}$~\cite{changedetection-EGRCNN}        & {15.7M}   & {61.1}  & UNet~\cite{UNet}                               & 84.36           & 78.41           & 81.64          & 69.02          & 91.64          \\
  MSPSNet$_{22}$~\cite{changedetection-MSPSNet}      & {2.2M}    & {132.7} & MSSNN~\cite{changedetection-MSPSNet}           & 78.47           & 74.91           & 77.39          & 63.13          & 89.43          \\
  ICIFNet$_{22}$~\cite{changedetection-ICIF-Net}     & {25.8M}   & {27.0}  & ResNet-18~\cite{Resnet}+PVT-V2-B1~\cite{PVTv2} & 83.37           & \third{78.51}   & 80.74          & 68.12          & 91.24          \\
  FTN$_{22}$~\cite{changedetection-FTN}              & {168.5M}  & {24.5}  & Swin-B~\cite{Swin}                             & 86.86           & 76.82           & 81.53          & 68.82          & 91.79          \\
  TransY-Net$_{23}$~\cite{changedetection-TransYNet} & {168.5M}  & {24.5}  & Swin-B~\cite{Swin}                             & \second{89.09}  & 77.42           & \third{82.84}  & \third{70.71}  & \third{92.44}  \\
  DMINet$_{23}$~\cite{changedetection-DMINet}        & {6.8M}    & {141.1} & ResNet-18~\cite{Resnet}                        & 85.03           & \best{79.86}    & 82.08          & 69.60          & 91.67          \\
  \midrule[0.5pt]
  SwinNet$^{\daleth}_{22}$~\cite{rgbdtsod-SwinNet}   & {199.2M}  & {26.0}  & Swin-B~\cite{Swin}                             & 87.28           & 77.19           & 81.93          & 69.38          & 91.97          \\
  DFormer-L$^{\daleth}_{24}$~\cite{mmseg-DFormer}    & {39.0M}   & {35.5}  & DFormer-L~\cite{mmseg-DFormer}                 & 87.02           & 76.87           & 81.63          & 68.96          & 91.84          \\
  \midrule[0.5pt]
  ComPtr-T                                           & {38.3M}   & {30.8}  & Swin-T~\cite{Swin}                             & \third{87.46}   & \second{79.75}  & \best{83.43}   & \best{71.57}   & \second{92.53} \\
  ComPtr-B                                           & {105.7M}  & {21.6}  & Swin-B~\cite{Swin}                             & \best{89.62}    & 77.61           & \second{83.18} & \second{71.21} & \best{92.60}   \\
  \bottomrule[2pt]
\end{tabular}

    }
    \captionof{table}{Comparison on SYSU-CD~\cite{changedetection-SYSU} for RSCD.}
    \label{tab:rscd-sysu}
  \end{minipage}
  \vspace{-1em}
\end{figure}
\begin{figure}[!t]
  \begin{minipage}{\linewidth}
    \centering
    \resizebox{\linewidth}{!}{%
      \rowcolors{2}{gray!10}{white}
      \begin{tabular}{l|rr|l|*5{c}}
  \toprule[2pt]
  Methods                                          & {Params.} & {FPS}  & Backbone                       & GAME$_0$~$\downarrow$ & GAME$_1$~$\downarrow$ & GAME$_2$~$\downarrow$ & GAME$_3$~$\downarrow$ & RMSE~$\downarrow$ \\
  \midrule[1pt]
  CSRNet$_{18}$~\cite{cc-CSRNet}                   & {16.3M}   & {91.2} & VGG-16~\cite{VGG}              & 20.40                 & 23.58                 & 28.03                 & 35.51                 & 35.26             \\
  BL$_{19}$~\cite{cc-BL}                           & {21.5M}   & {88.5} & VGG-19~\cite{VGG}              & 18.70                 & 22.55                 & 26.83                 & 34.62                 & 32.67             \\
  DM-Count$_{20}$~\cite{cc-dm}                     & {21.5M}   & {89.0} & VGG-19~\cite{VGG}              & 16.54                 & 20.73                 & 25.23                 & 32.23                 & 27.22             \\
  P2PNet$_{21}$~\cite{cc-P2PNet}                   & {21.6M}   & {96.8} & VGG-16~\cite{VGG}              & 16.24                 & 19.42                 & 23.48                 & 30.27                 & 29.94             \\
  MARUNet$_{21}$~\cite{cc-MARUNet-CFANet}          & {27.3M}   & {17.8} & VGG-16~\cite{VGG}              & 17.39                 & 20.54                 & 23.69                 & 27.36                 & 30.84             \\
  MAN$_{22}$~\cite{cc-MAN}                         & {40.4M}   & {71.1} & VGG-19~\cite{VGG}              & 17.16                 & 21.78                 & 28.74                 & 41.59                 & 33.84             \\
  \midrule[0.5pt]
  CMCRL$_{21}$~\cite{rgbtcc-CMCRL-RGBTCC}          & {25.7M}   & {38.1} & VGG-16~\cite{VGG}              & 15.61                 & 19.95                 & 24.69                 & 32.89                 & 28.18             \\
  TAFNet$_{22}$~\cite{rgbtcc-TAFNet}               & {N/A}     & {N/A}  & VGG-16~\cite{VGG}              & 12.38                 & 16.98                 & 21.86                 & 30.19                 & 22.45             \\
  MAT$_{22}$~\cite{rgbtcc-MAT}                     & {N/A}     & {N/A}  & VGG-16~\cite{VGG}              & 12.35                 & 16.29                 & 20.81                 & 29.09                 & 22.53             \\
  DEFNet$_{22}$~\cite{rgbtcc-DEFNet}               & {45.3M}   & {15.3} & VGG-16~\cite{VGG}              & 11.90                 & 16.08                 & 20.19                 & 27.27                 & 21.09             \\
  MMCC$_{22}$~\cite{rgbtcc-RGBTMMCC}               & {261.9M}  & {19.7} & PVT-V2-B3~\cite{PVTv2}         & \second{10.90}        & \second{14.81}        & \second{19.02}        & \second{26.14}        & \second{18.79}    \\
  CCANet$_{24}$~\cite{rgbtcc-CCANet}               & {N/A}     & {N/A}  & VGG-16~\cite{VGG}              & 13.93                 & 18.13                 & 22.08                 & 28.26                 & 24.71             \\
  \midrule[0.5pt]
  SwinNet$^{\daleth}_{23}$~\cite{rgbdtsod-SwinNet} & {199.2M}  & {11.4} & Swin-B~\cite{Swin}             & 11.95                 & 15.97                 & 20.87                 & 26.48                 & 21.15             \\
  DFormer-L$^{\daleth}_{24}$~\cite{mmseg-DFormer}  & {39.0M}   & {26.8} & DFormer-L~\cite{mmseg-DFormer} & 12.13                 & 16.14                 & 21.08                 & 26.84                 & 21.78             \\
  \midrule[0.5pt]
  ComPtr-T                                         & {38.3M}   & {29.6} & Swin-T~\cite{Swin}             & \best{10.52}          & \best{14.51}          & \best{18.48}          & \best{24.28}          & \best{18.48}      \\
  ComPtr-B                                         & {105.7M}  & {21.1} & Swin-B~\cite{Swin}             & \third{11.82}         & \third{15.84}         & \third{20.17}         & \third{26.41}         & \third{20.75}     \\
  \bottomrule[2pt]
\end{tabular}

    }
    \captionof{table}{Comparison on RGBT-CC~\cite{rgbtcc-CMCRL-RGBTCC}.
      The top six single-modal methods are retrained by~\cite{rgbtcc-RGBTMMCC} and the others are bi-modal methods.
    }
    \label{tab:rgbtcc}
  \end{minipage}
  \par
  \begin{minipage}{\linewidth}
    \vspace{1em}
    \centering
    \resizebox{0.7\linewidth}{!}{%
      \rowcolors{2}{gray!10}{white}
      \begin{tabular}{l|rr|l|*3{c}}
  \toprule[2pt]
  Methods                                               & {Params.} & {FPS}  & Backbone                                           & mIoU~$\uparrow$ \\
  \midrule[1pt]
  3DGNN$_{17}$~\cite{mmseg-3DGNN}                       & {N/A}     & {N/A}  & VGG-16~\cite{VGG}                                  & 45.9            \\
  RDF$_{17}$~\cite{mmseg-RDFNet}                        & {443.8M}  & {11.4} & ResNet-152~\cite{Resnet}                           & 47.7            \\
  ACNet$_{19}$~\cite{mmseg-ACNet}                       & {116.6M}  & {21.2} & ResNet-50~\cite{Resnet}                            & 49.4            \\
  CEN$_{20}$~\cite{mmseg-CEN}                           & {133.9M}  & {10.1} & ResNet-152~\cite{Resnet}                           & 51.1            \\
  SA-Gate$_{20}$~\cite{mmseg-SA-Gate}                   & {110.6M}  & {16.6} & ResNet-101~\cite{Resnet}                           & 49.4            \\
  SGNet$_{21}$~\cite{mmseg-SGNet}                       & {58.3M}   & {24.0} & ResNet-101~\cite{Resnet}                           & 48.6            \\
  NANet$_{21}$~\cite{mmseg-NANet}                       & {N/A}     & {N/A}  & ResNet-101~\cite{Resnet}                           & 48.8            \\
  ShapeConv$_{21}$~\cite{mmseg-ShapeConv}               & {110.0M}  & {29.2} & ResNeXt-101~\cite{ResNext}                         & 48.6            \\
  FSFNet$_{21}$~\cite{mmseg-FSFNet}                     & {N/A}     & {N/A}  & ResNet-101~\cite{Resnet}                           & 50.6            \\
  FRNet$_{22}$~\cite{mmseg-FRNet}                       & {87.8M}   & {54.2} & ResNet-34~\cite{Resnet}                            & 51.8            \\
  EMSANet$_{22}$~\cite{mmseg-EMSANet}                   & {46.9M}   & {56.0} & ResNet-34~\cite{Resnet}                            & 48.2            \\
  TokenFusion$^{\dagger}_{22}$~\cite{mmseg-TokenFusion} & {45.9M}   & {14.9} & MiT-B3~\cite{Segformer}                            & 51.0            \\
  MultiMAE$^{\dagger}_{22}$~\cite{mmseg-MultiMAE}       & {97.8M}   & {7.1}  & ViT-B~\cite{ViT}                                   & 51.1            \\
  CMNeXt$^{\dagger}_{23}$~\cite{mmseg-CMNeXt}           & {116.6M}  & {16.1} & MiT-B4~\cite{Segformer}                            & 51.9            \\
  PGDENet$_{23}$~\cite{mmseg-PGDENet}                   & {100.7M}  & {17.2} & ResNet-34~\cite{Resnet}                            & 51.0            \\
  CMX$_{23}$~\cite{mmseg-CMX}                           & {181.1M}  & {12.8} & MiT-B5~\cite{Segformer}                            & \third{52.4}    \\
  EMSAFormer$_{23}$~\cite{mmseg-EMSAFormer}             & {52.6M}   & {63.4} & SwinV2-T~\cite{SwinV2}                             & 48.8            \\
  AsymFormer$_{23}$~\cite{mmseg-Asymformer}             & {33.1M}   & {63.8} & ConvNeXt-T~\cite{ConvNeXt}+MiT-B0~\cite{Segformer} & 49.1            \\
  DFormer-L$_{24}$~\cite{mmseg-DFormer}                 & {39.0M}   & {29.0} & DFormer-L~\cite{mmseg-DFormer}                     & \second{52.5}   \\
  \midrule[0.5pt]
  SwinNet$^{\daleth}_{22}$~\cite{rgbdtsod-SwinNet}      & {199.2M}  & {18.7} & Swin-B~\cite{Swin}                                 & 51.7            \\
  \midrule[0.5pt]
  ComPtr-T                                              & {38.3M}   & {27.1} & Swin-T~\cite{Swin}                                 & 48.9            \\
  ComPtr-B                                              & {105.7M}  & {20.3} & Swin-B~\cite{Swin}                                 & \best{52.8}     \\
  \bottomrule[2pt]
\end{tabular}

    }
    \captionof{table}{Comparison on SUN-RGBD~\cite{SUN-RGBD-37classes} for RGB-D SS.
      $\dagger$ indicates that we follow the results from DFormer~\cite{mmseg-DFormer}.}
    \label{tab:rgbdss}
  \end{minipage}
  \vspace{-1em}
\end{figure}

\begin{figure*}[!t]
  \begin{minipage}{\linewidth}
    \centering
    \resizebox{\linewidth}{!}{%
      \rowcolors{2}{gray!10}{white}
      \begin{tabular}{l|rr|l*{5}{|*{5}{c}}}
  \toprule[2pt]
                                                  &
                                                  &
                                                  &
                                                  &
  \multicolumn{5}{c}{DUTLF-Depth~\cite{DUTRGBD}}  &
  \multicolumn{5}{|c}{NJUD~\cite{NJUD}}           &
  \multicolumn{5}{|c}{NLPR~\cite{NLPR}}           &
  \multicolumn{5}{|c}{SIP~\cite{SIP}}             &
  \multicolumn{5}{|c}{STEREO1000~\cite{STEREO}}                                                                                                                                                                                                                                                                                                                                                                                                                                                                                                                                                                                                                                                                                                                 \\
  \multirow{-2}{*}{Methods}                       & {\multirow{-2}{*}{Params.}} & {\multirow{-2}{*}{FPS}} & \multirow{-2}{*}{Backbone}               & $S_m$~$\uparrow$ & $F^{\omega}_{\beta}$~$\uparrow$ & MAE~$\downarrow$ & $E_m$~$\uparrow$ & $F_{\beta}$~$\uparrow$ & $S_m$~$\uparrow$ & $F^{\omega}_{\beta}$~$\uparrow$ & MAE~$\downarrow$ & $E_m$~$\uparrow$ & $F_{\beta}$~$\uparrow$ & $S_m$~$\uparrow$ & $F^{\omega}_{\beta}$~$\uparrow$ & MAE~$\downarrow$ & $E_m$~$\uparrow$ & $F_{\beta}$~$\uparrow$ & $S_m$~$\uparrow$ & $F^{\omega}_{\beta}$~$\uparrow$ & MAE~$\downarrow$ & $E_m$~$\uparrow$ & $F_{\beta}$~$\uparrow$ & $S_m$~$\uparrow$ & $F^{\omega}_{\beta}$~$\uparrow$ & MAE~$\downarrow$ & $E_m$~$\uparrow$ & $F_{\beta}$~$\uparrow$ \\
  \midrule[1pt]
  JLDCF$_{20}$~\cite{JLDCF-RGBDSOD-journal}       & {143.5M}                    & {32.6}                  & ResNet-101~\cite{Resnet}                 & 90.52            & 86.33                           & 4.29             & 94.26            & 91.12                  & 90.22            & 86.91                           & 4.13             & 94.37            & 90.42                  & 92.50            & 88.20                           & 2.16             & 96.27            & 91.76                  & 88.04            & 84.38                           & 4.91             & 92.47            & 88.90                  & 90.26            & 85.70                           & 4.03             & 94.68            & 90.39                  \\
  HDFNet$_{20}$~\cite{HDFNet}                     & {153.2M}                    & {52.1}                  & ResNet-50~\cite{Resnet}                  & 90.72            & 86.40                           & 4.13             & 94.73            & 91.76                  & 90.79            & 87.72                           & 3.81             & 94.42            & 91.07                  & 92.32            & 88.22                           & 2.28             & 96.29            & 91.71                  & 88.59            & 84.77                           & 4.72             & 92.97            & 89.38                  & 89.97            & 85.26                           & 4.13             & 94.34            & 89.96                  \\
  D3Net$_{20}$~\cite{SIP}                         & {45.2M}                     & {127.2}                 & VGG-19~\cite{VGG}                        & 77.53            & 66.80                           & 9.67             & 83.30            & 74.17                  & 90.05            & 85.41                           & 4.63             & 93.85            & 90.00                  & 91.17            & 84.86                           & 2.96             & 95.30            & 89.69                  & 86.03            & 79.89                           & 6.31             & 90.86            & 86.11                  & 89.86            & 83.78                           & 4.58             & 93.83            & 89.12                  \\
  RD3D$_{21}$~\cite{RD3D-RGBDSOD}                 & {46.9M}                     & {25.9}                  & I3DResNet-50~\cite{ResNet-I3D}           & 93.13            & 90.74                           & 3.11             & 95.97            & 93.87                  & 91.56            & 88.55                           & 3.65             & 94.71            & 91.39                  & 92.96            & 88.87                           & 2.21             & 96.49            & 91.90                  & 88.53            & 84.49                           & 4.84             & 92.43            & 88.90                  & 91.13            & 87.06                           & 3.74             & 94.66            & 90.61                  \\
  TriTransNet$_{21}$~\cite{TriTransNet-RGBDSOD}   & {139.5M}                    & {10.2}                  & ResNet-50~\cite{Resnet}+ViT-B~\cite{ViT} & 93.26            & {92.63}                         & {2.49}           & 96.57            & 94.63                  & 91.96            & 90.59                           & 3.03             & 95.50            & 92.63                  & 92.84            & 90.18                           & {2.04}           & 96.62            & 92.38                  & 88.61            & 86.42                           & 4.34             & 93.03            & 89.88                  & 90.78            & 88.16                           & 3.33             & 95.32            & 91.14                  \\
  DCF$_{21}$~\cite{DCF-RGBDSOD}                   & {108.5M}                    & {23.9}                  & ResNet-50~\cite{Resnet}                  & 92.40            & 90.94                           & 3.01             & 95.71            & 93.19                  & 90.37            & 87.65                           & 3.87             & 94.31            & 90.52                  & 92.20            & 88.41                           & 2.35             & 95.74            & 90.99                  & 87.35            & 83.97                           & 5.19             & 92.18            & 88.59                  & 90.57            & 87.19                           & 3.70             & 94.77            & 90.44                  \\
  DSA2F$_{21}$~\cite{DSA2F-RGBDSOD}               & {36.5M}                     & {25.4}                  & VGG-19~\cite{VGG}                        & 92.18            & 90.73                           & 3.09             & 95.81            & 93.26                  & 90.43            & 88.19                           & 3.94             & 93.87            & 90.74                  & 91.88            & 88.00                           & 2.42             & 95.23            & 90.59                  & 86.20            & 82.77                           & 5.74             & 91.21            & 87.54                  & 89.78            & 86.79                           & 3.87             & 94.20            & 89.96                  \\
  CMINet$^{\star}_{21}$~\cite{COME15K-CMINet}     & {185.5M}                    & {10.3}                  & ResNet-50~\cite{Resnet}                  & 89.68            & 86.66                           & 4.61             & 92.58            & 89.09                  & 92.86            & {90.95}                         & 2.86             & 95.74            & {93.36}                & 93.17            & 90.01                           & 2.05             & 96.24            & 92.23                  & 89.85            & 87.17                           & 4.04             & 93.89            & 90.99                  & 91.83            & {88.55}                         & {3.23}           & 95.12            & {91.64}                \\
  SPNet$_{21}$~\cite{SPNet-RGBDSOD-journal}       & {175.3M}                    & {26.9}                  & Res2Net-50~\cite{Res2Net}                & 80.41            & 73.46                           & 8.53             & 90.98            & 84.89                  & 92.45            & 90.61                           & {2.85}           & 95.72            & 92.80                  & 92.73            & 89.60                           & 2.08             & 96.16            & 91.85                  & 89.37            & 86.75                           & 4.30             & 93.28            & 90.36                  & 90.68            & 87.31                           & 3.69             & 94.87            & 90.63                  \\
  VST$_{21}$~\cite{rgbdsod-VST}                   & {84.0M}                     & {32.9}                  & T2T-ViT-t14~\cite{T2T-ViT}               & {94.25}          & 92.05                           & {2.50}           & {97.08}          & {94.90}                & 92.24            & 88.82                           & 3.43             & 95.10            & 91.95                  & 93.14            & 88.66                           & 2.33             & 96.23            & 92.01                  & 90.36            & 87.25                           & 3.96             & 94.39            & 91.50                  & 91.28            & 86.56                           & 3.76             & 95.06            & 90.67                  \\
  HAINet$_{21}$~\cite{HAINet-RGBDSOD}             & {59.8M}                     & {12.2}                  & VGG-16~\cite{VGG}                        & 90.95            & 88.32                           & 3.81             & 94.42            & 91.98                  & 90.93            & 87.86                           & 3.84             & 94.05            & 90.95                  & 92.10            & 88.07                           & 2.50             & 95.38            & 90.79                  & 88.61            & 85.42                           & 4.84             & 92.71            & 90.26                  & 90.93            & 87.14                           & 3.80             & 94.71            & 90.93                  \\
  CCAFNet$_{21}$~\cite{CCAFNet-RGBDSOD}           & {41.8M}                     & {64.2}                  & VGG-16~\cite{VGG}                        & 90.40            & 87.81                           & 3.77             & 94.31            & 91.32                  & 91.01            & 87.65                           & 3.74             & 94.40            & 90.99                  & 92.21            & 87.48                           & 2.67             & 95.67            & 90.87                  & 87.69            & 82.93                           & 5.44             & 91.65            & 88.04                  & 89.17            & 84.40                           & 4.46             & 93.42            & 88.65                  \\
  DCMF$_{22}$~\cite{DCMF-RGBDSOD}                 & {58.9M}                     & {23.1}                  & VGG-16~\cite{VGG}                        & 92.79            & 88.85                           & 3.50             & 95.82            & 93.19                  & 91.26            & 86.69                           & 4.26             & 94.79            & 91.54                  & 92.20            & 85.62                           & 2.87             & 95.41            & 90.56                  & 87.00            & 80.84                           & 6.18             & 91.10            & 87.19                  & 90.97            & 84.87                           & 4.26             & 94.64            & 90.63                  \\
  SwinNet$_{22}$~\cite{rgbdtsod-SwinNet}          & {199.2M}                    & {25.8}                  & Swin-B~\cite{Swin}                       & 93.36            & 91.28                           & 2.90             & 96.25            & 94.04                  & \second{93.47}   & {91.27}                         & 2.71             & \third{96.33}    & \third{93.82}          & \third{94.13}    & {90.84}                         & \second{1.79}    & \best{97.36}     & \best{93.61}           & \third{91.13}    & \third{89.00}                   & {3.52}           & \third{94.98}    & \third{92.70}          & {91.93}          & 88.21                           & 3.28             & \third{95.60}    & \third{91.78}          \\
  CAVER$_{23}$~\cite{caver-tip}                   & {93.8M}                     & {27.7}                  & ResNet-101D~\cite{Resnet}                & 93.74            & {92.58}                         & 2.59             & 96.75            & 94.63                  & 92.55            & 90.60                           & 2.89             & {95.88}          & 92.83                  & 93.41            & {90.42}                         & 2.08             & \third{97.00}    & {92.90}                & 90.39            & 87.87                           & 3.70             & 94.47            & 91.51                  & 91.74            & {88.78}                         & {3.19}           & {95.52}          & 91.61                  \\
  HRTransNet$_{23}$~\cite{rgbdtsod-HRTransNet}    & {73.6M}                     & {14.6}                  & HRFormer-B~\cite{HRFormer}               & \second{95.07}   & \second{94.93}                  & \second{1.80}    & \second{97.76}   & \second{96.22}         & 93.33            & \second{92.00}                  & \second{2.57}    & \second{96.32}   & \second{93.92}         & \second{94.18}   & \best{91.70}                    & \best{1.60}      & \best{97.36}     & \second{93.59}         & 90.90            & \best{89.56}                    & \second{3.46}    & 94.94            & \second{92.88}         & \third{92.12}    & \second{89.37}                  & \third{3.01}     & \second{95.63}   & \second{91.91}         \\
  \midrule[0.5pt]
  DFormer-L$^{\daleth}_{24}$~\cite{mmseg-DFormer} & {39.0M}                     & {36.0}                  & DFormer-L~\cite{mmseg-DFormer}           & \third{94.62}    & \third{94.24}                   & \third{2.21}     & \third{97.20}    & \third{95.71}          & \third{93.43}    & \best{92.03}                    & \third{2.58}     & 96.30            & 93.79                  & 94.02            & \second{90.97}                  & 1.95             & 97.04            & 93.31                  & \third{91.13}    & 88.95                           & \third{3.51}     & 94.94            & 92.61                  & \second{92.31}   & \third{89.20}                   & \second{2.99}    & \third{95.60}    & 91.58                  \\
  \midrule[0.5pt]
  ComPtr-T                                        & {38.3M}                     & {30.8}                  & Swin-T~\cite{Swin}                       & 94.37            & 92.17                           & {2.50}           & {96.99}          & {95.06}                & 92.89            & 89.48                           & 3.27             & 95.84            & 92.99                  & {93.56}          & 88.66                           & 2.14             & 96.68            & 92.40                  & \second{91.34}   & {87.96}                         & {3.61}           & \second{94.99}   & {92.11}                & {91.99}          & 87.45                           & 3.48             & 95.47            & 91.55                  \\
  ComPtr-B                                        & {105.7M}                    & {22.9}                  & Swin-B~\cite{Swin}                       & \best{95.87}     & \best{95.14}                    & \best{1.69}      & \best{98.19}     & \best{96.80}           & \best{93.96}     & \third{91.95}                   & \best{2.49}      & \best{96.64}     & \best{94.46}           & \best{94.29}     & \third{90.73}                   & \third{1.94}     & \second{97.11}   & \third{93.57}          & \best{91.53}     & \second{89.48}                  & \best{3.37}      & \best{95.04}     & \best{93.07}           & \best{93.29}     & \best{90.29}                    & \best{2.79}      & \best{96.23}     & \best{93.24}           \\
  \bottomrule[2pt]
\end{tabular}

    }
    \captionof{table}{Experiments on five RGB-D SOD benchmarks.
      ``$\star$'': Using the multi-scale training strategy.
    }
    \label{tab:rgbdsod}
  \end{minipage}
  \par
  \begin{minipage}{\linewidth}
    \vspace{1em}
    \centering
    \resizebox{0.7\linewidth}{!}{%
      \rowcolors{2}{gray!10}{white}
      \begin{tabular}{l|rr|l|*5{c}|*5{c}|*5{c}}
  \toprule[2pt]
                                                  &
                                                  &
                                                  &
                                                  &
  \multicolumn{5}{c}{VT821~\cite{VT821-MTMR}}     &
  \multicolumn{5}{|c}{VT1000~\cite{VT1000-SDGL}}  &
  \multicolumn{5}{|c}{VT5000~\cite{VT5000-ADF}}                                                                                                                                                                                                                                                                                                                                                                                                                                                                      \\
  \multirow{-2}{*}{Methods}                       & {\multirow{-2}{*}{Params.}} & {\multirow{-2}{*}{FPS}} & \multirow{-2}{*}{Backbone}     & $S_m$~$\uparrow$ & $F^{\omega}_{\beta}$~$\uparrow$ & MAE~$\downarrow$ & $E_m$~$\uparrow$ & $F_{\beta}$~$\uparrow$ & $S_m$~$\uparrow$ & $F^{\omega}_{\beta}$~$\uparrow$ & MAE~$\downarrow$ & $E_m$~$\uparrow$ & $F_{\beta}$~$\uparrow$ & $S_m$~$\uparrow$ & $F^{\omega}_{\beta}$~$\uparrow$ & MAE~$\downarrow$ & $E_m$~$\uparrow$ & $F_{\beta}$~$\uparrow$ \\
  \midrule[1pt]
  MTMR$_{18}$~\cite{VT821-MTMR}                   & {N/A}                       & {N/A}                   & \blank                         & 59.27            & 26.39                           & 25.95            & 81.08            & 64.64                  & 70.57            & 48.54                           & 11.94            & 83.59            & 71.54                  & 68.01            & 39.68                           & 11.43            & 79.23            & 61.33                  \\
  M3S-NIR$_{19}$~\cite{M3S-NIR-RGBTSOD}           & {N/A}                       & {N/A}                   & \blank                         & 72.31            & 40.72                           & 13.97            & 83.71            & 73.78                  & 72.58            & 46.28                           & 14.54            & 82.79            & 73.53                  & 65.21            & 32.71                           & 16.80            & 75.99            & 59.57                  \\
  ADF$_{20}$~\cite{VT5000-ADF}                    & {N/A}                       & {N/A}                   & \blank                         & 81.02            & 62.65                           & 7.65             & 83.89            & 75.22                  & 90.95            & 80.43                           & 3.39             & 95.05            & 90.81                  & 86.35            & 72.18                           & 4.82             & 91.11            & 83.68                  \\
  SDGL$_{20}$~\cite{VT1000-SDGL}                  & {N/A}                       & {N/A}                   & VGG-19~\cite{VGG}              & 76.54            & 58.28                           & 8.49             & 83.95            & 73.47                  & 78.67            & 65.24                           & 8.96             & 85.85            & 77.00                  & 75.04            & 55.85                           & 8.86             & 82.92            & 69.49                  \\
  CSRNet$_{21}$~\cite{CSRNet-RGBTSOD}             & {1.0M}                      & {43.1}                  & ESPNetv2~\cite{ESPNetv2}       & 88.47            & 82.12                           & 3.76             & 92.26            & 85.79                  & 91.83            & 87.82                           & 2.42             & 95.26            & 90.83                  & 86.75            & 79.64                           & 4.17             & 91.38            & 83.70                  \\
  ECFFNet$_{21}$~\cite{ECFFNet-RGBTSOD}           & {N/A}                       & {N/A}                   & ResNet-34~\cite{Resnet}        & 87.71            & 79.93                           & 3.47             & 91.03            & 83.52                  & 92.38            & 88.35                           & 2.17             & 95.93            & 91.71                  & 87.45            & 79.98                           & 3.79             & 92.12            & 84.61                  \\
  MIDD$_{21}$~\cite{MIDD-RGBTSOD}                 & {52.4M}                     & {21.5}                  & VGG-16~\cite{VGG}              & 87.10            & 75.96                           & 4.45             & 91.83            & 85.08                  & 91.54            & 85.58                           & 2.71             & 95.67            & 91.28                  & 86.74            & 76.29                           & 4.32             & 92.01            & 84.91                  \\
  CGFNet$_{21}$~\cite{CGFNet-RGBTSOD}             & {66.4M}                     & {12.5}                  & VGG-16~\cite{VGG}              & 88.05            & 82.87                           & 3.78             & 92.04            & 86.58                  & 92.32            & 89.99                           & 2.31             & 95.94            & 92.35                  & 88.30            & 83.11                           & 3.53             & 92.65            & 86.90                  \\
  SwinNet$_{22}$~\cite{rgbdtsod-SwinNet}          & {199.2M}                    & {25.8}                  & Swin-B~\cite{Swin}             & 90.37            & 81.83                           & 2.98             & \third{93.87}    & 88.09                  & \third{93.81}    & 89.36                           & 1.79             & \third{97.43}    & \third{94.09}          & \third{91.17}    & 84.61                           & \third{2.58}     & \third{95.38}    & \third{90.18}          \\
  CAVER$_{23}$~\cite{caver-tip}                   & {93.8M}                     & {27.7}                  & ResNet-101D~\cite{ResNet-D}    & 89.78            & \third{84.62}                   & \third{2.64}     & 93.40            & 87.67                  & 93.76            & \third{91.24}                   & \second{1.64}    & 97.28            & 93.93                  & 89.94            & {84.89}                         & 2.84             & 94.39            & 88.18                  \\
  HRTransNet$_{23}$~\cite{rgbdtsod-HRTransNet}    & {73.6M}                     & {14.6}                  & HRFormer-B~\cite{HRFormer}     & \second{90.59}   & \second{84.92}                  & \second{2.57}    & \second{94.11}   & \second{88.85}         & 93.79            & \second{91.27}                  & {1.71}           & \second{97.50}   & 94.07                  & \second{91.23}   & \second{86.99}                  & \second{2.53}    & \second{95.59}   & \second{90.30}         \\
  \midrule[0.5pt]
  DFormer-L$^{\daleth}_{24}$~\cite{mmseg-DFormer} & {39.0M}                     & {36.0}                  & DFormer-L~\cite{mmseg-DFormer} & 90.22            & 84.25                           & 2.70             & 93.82            & 88.20                  & 93.75            & 91.22                           & \third{1.69}     & 97.33            & 93.74                  & 90.87            & \third{85.51}                   & 2.78             & 94.81            & 89.56                  \\
  \midrule[0.5pt]
  ComPtr-T                                        & {38.3M}                     & {30.8}                  & Swin-T~\cite{Swin}             & \third{90.52}    & 83.32                           & 2.91             & 93.79            & \third{88.50}          & \second{94.17}   & 89.82                           & 1.95             & 97.41            & \second{94.21}         & 90.71            & 83.10                           & 3.11             & 94.71            & 88.80                  \\
  ComPtr-B                                        & {105.7M}                    & {22.9}                  & Swin-B~\cite{Swin}             & \best{92.37}     & \best{87.48}                    & \best{2.16}      & \best{95.01}     & \best{90.65}           & \best{95.12}     & \best{92.66}                    & \best{1.47}      & \best{98.03}     & \best{95.36}           & \best{92.84}     & \best{88.34}                    & \best{2.21}      & \best{96.29}     & \best{91.86}           \\
  \bottomrule[2pt]
\end{tabular}

    }
    \captionof{table}{Comparison on three RGB-T SOD benchmarks.
      ``\blank'': Traditional methods based on hand-crafted features.
    }
    \label{tab:rgbtsod}
  \end{minipage}
  \vspace{-1em}
\end{figure*}

\subsection{RGB-T Crowd Counting (RGB-T CC)}\label{sec:rgbtcc-setting}

\noindent\textbf{Datasets.}
In the pioneering work~\cite{rgbtcc-CMCRL-RGBTCC}, a new large-scale point-annotated dataset benchmark \textbf{RGBT-CC} is carefully collected and manually aligned by an optical-thermal camera and contains 2030 RGB-T pairs with 138389 point-annotated people.
All samples have a standard resolution of $640 \times 480$.
1013 pairs are captured in the light and 1017 pairs are taken from some dark scenes.
The number of the training, validation and testing sets is 1030, 200 and 800, respectively.
The model is trained, validated, and tested independently on these datasets like the existing methods~\cite{changedetection-DMINet,rgbtcc-RGBTMMCC}.
All settings are aligned with the recent best practice~\cite{rgbtcc-RGBTMMCC}, where the batch size, learning rate, weight decay, epoch, and crop size are set to 16, 0.00001, 0.0001, 500, $224 \times 224$.
And no data augmentation other than random cropping is used for this task.

\noindent\textbf{Metrics.}
For RGB-T CC, five metrics are utilized to evaluate the regression error, {including} four levels of \textit{grid average mean absolute error}~\cite{cc-GAME} (GAME$_i$, $i \in \{0, 1, 2, 3\}$) and \textit{root mean square error} (RMSE).
To calculate GAME, the density prediction and the ground truth (GT) are grided into $4^i$ ($i \in \{0, 1, 2, 3\}$) parts: $\mathrm{GAME}_{i} = \frac{1}{N} \sum^{N}_{n=1} \sum^{4^i}_{j=1} | P^j_n - G^j_n |$,
where $P^j_n$ and $G^j_n$ are the count prediction and GT in $j^{th}$ part of $n^{th}$ test image pairs.
RMSE is mathematically defined as: $\mathrm{RMSE} = \sqrt{\frac{1}{N} \sum^{N}_{n=1} (P_n - G_n)^2}$,
where $P_n$ and $G_n$ are the predicted count and GT of $n^{th}$ test image pairs.

\subsection{RGB-D/T Salient Object Detection (RGB-D/T SOD)}\label{sec:rgbdtsod-setting}

\noindent\textbf{Datasets.}
We introduce 5 datasets including NJUD~\cite{NJUD}, NLPR~\cite{NLPR}, STEREO1000~\cite{STEREO}, SIP~\cite{SIP}, and DUTLF-Depth~\cite{DUTRGBD} for RGB-D SOD,
and 3 datasets including VT821~\cite{VT821-MTMR}, VT1000~\cite{VT1000-SDGL}, and VT5000~\cite{VT5000-ADF} for RGB-T SOD.
\textbf{NJUD} consists of 1985 image pairs involving a lot of complex scenarios.
In \textbf{NLPR}, 1000 RGB-D image pairs are collected from diverse indoor and outdoor scenes.
\textbf{STEREO1000} is composed of 1000 RGB-D image pairs from Flickr, NVIDIA 3D Vision Live and Stereoscopic Image Gallery.
\textbf{SIP} contains 929 pairs of high-resolution person images captured in challenging outdoor scenes.
Larger-scale \textbf{DUTLF-Depth} collects 800 pairs of indoor and 400 pairs of outdoor RGB-D images, which contain a large number of challenging objects and scenes.
Continuing the setup of existing methods~\cite{caver-tip,HAINet-RGBDSOD,TriTransNet-RGBDSOD}, we directly use the same 2985 images from NJUD, NLPR, and DUTLF-Depth to train the model and evaluate it on the remaining data.
\textbf{VT821} is a pioneering RGB-T SOD {dataset}, which includes 821 RGB-Thermal-GT image pairs.
\textbf{VT1000} containing 1000 pairs of images, increases the data scale and covers more than 400 kinds of objects in more than 10 types of scenes with diverse lighting conditions.
\textbf{VT5000} contains 5000 pairs of densely annotated RGB-T images, which greatly enriches the diversity and complexity of the task.
Except for the training set containing 2500 image pairs of VT5000 for training, the rest of the data are used for testing as~\cite{rgbdtsod-SwinNet,caver-tip}.
On RGB-D/T SOD, the batch size, learning rate, and input shape are consistently set to 16, 0.0002, and $256 \times 256$ for ComPtr-T as~\cite{caver-tip} ($384 \times 384$ for ComPtr-B as~\cite{rgbdtsod-SwinNet}).
Following~\cite{SPNet-RGBDSOD-journal,ECFFNet-RGBTSOD,rgbdtsod-SwinNet,caver-tip}, random affine transformation, flipping, color jittering are introduced as the data augmentation and our models are trained for 100 epochs in RGB-T SOD and 200 epochs in RGB-D SOD with the cosine scheduler.

\noindent\textbf{Metrics.}
For RGB-D/T SOD, all methods are evaluated based on five gray-scale image metrics,
\textit{S-measure}~\cite{Smeasure} ($S_m$),
\textit{maximum F-measure}~\cite{Fmeasure} ($F_{\beta}$),
\textit{maximum E-measure}~\cite{Emeasure} ($E_{m}$),
\textit{weighted F-measure}~\cite{wFmeasure} ($F^{\omega}_{\beta}$),
and \textit{MAE}.
$S_m$ focuses on region-aware and object-aware structural similarities $S_r$ and $S_o$, between the saliency map and the ground truth, which is expressed as: $S_m = 0.5 S_o + 0.5 S_r$.
$F_{\beta}$ is a region-based similarity metric based on precision and recall and its mathematical form is: $F_{\beta} = \frac{(1 + \beta^2) precision \times recall}{\beta^2 precision + recall}$, where $\beta^2$ is generally set to $0.3$ to emphasize more on the precision.
$E_m$ is composed of local pixel values and the image-level mean value to jointly evaluate the similarity between the prediction and the ground truth.
$F^{\omega}_{\beta}$ improves $F_{\beta}$ by using a weighted precision for measuring exactness and a weighted recall for measuring completeness.
MAE indicates the average absolute pixel error.
It is noteworthy that the evaluation of the SOD task is based on the gray-scale prediction map.
The value of the predicted result indicates the probability of the corresponding position belonging to the salient object region.
So, both the calculation process and the formula details of $F_{\beta}$ and $F1$ mentioned in Sec.~\ref{sec:rscd-setting} are different, which serves the specific task requirements.

\begin{figure*}[!t]
  \begin{minipage}{\linewidth}
    \centering
    \includegraphics[width=0.7\linewidth]{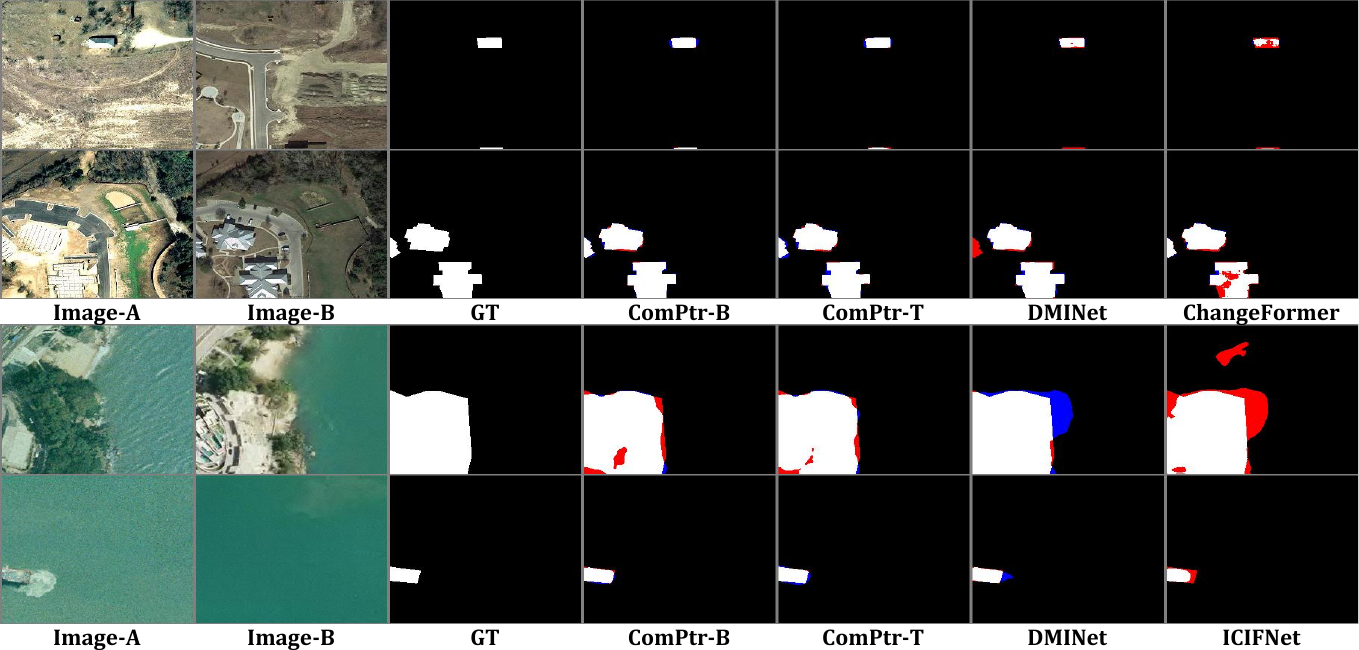}
    \captionof{figure}{Visual comparison on LEVIR-CD~\cite{changedetection-LEVIR} in the top two rows and SYSU-CD~\cite{changedetection-SYSU} in the bottom two rows for remote sensing change detection.
      \textcolor{reda}{Red} and \textcolor{myblue}{blue} colors in the binary prediction indicate false negative and false positive regions.
    }
    \label{fig:rscd-cmp}
  \end{minipage}
  \par
  \begin{minipage}{\linewidth}
    \vspace{1em}
    \centering
    \includegraphics[width=0.7\linewidth]{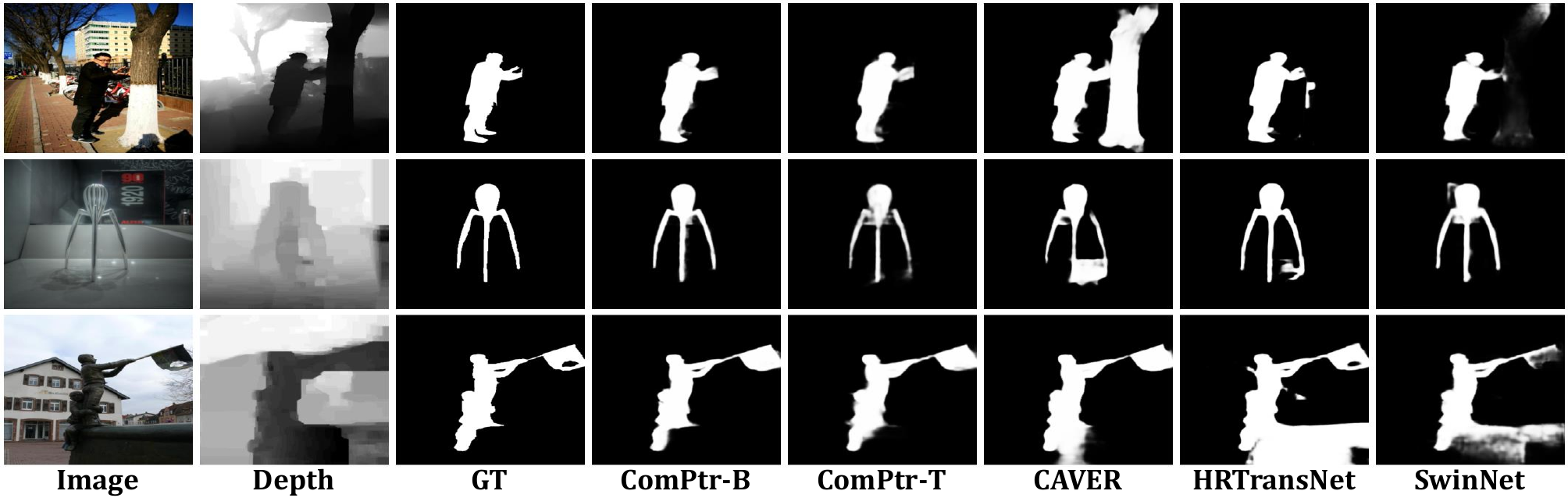}
    \captionof{figure}{Visual comparison for RGB-D salient object detection.}
    \label{fig:rgbdsod-cmp}
  \end{minipage}
  \par
  \begin{minipage}{\linewidth}
    \vspace{1em}
    \centering
    \includegraphics[width=0.7\linewidth]{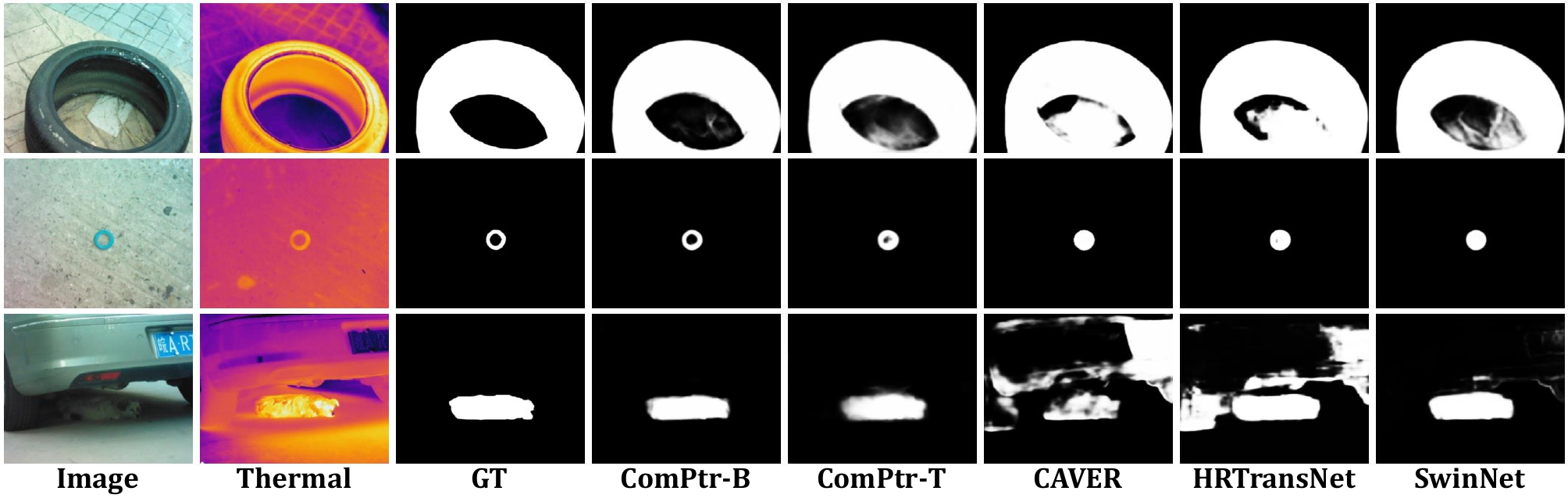}
    \captionof{figure}{Visual comparison for RGB-T salient object detection.}
    \label{fig:rgbtsod-cmp}
  \end{minipage}
\end{figure*}

\begin{figure*}[!t]
  \centering
  \includegraphics[width=0.7\linewidth]{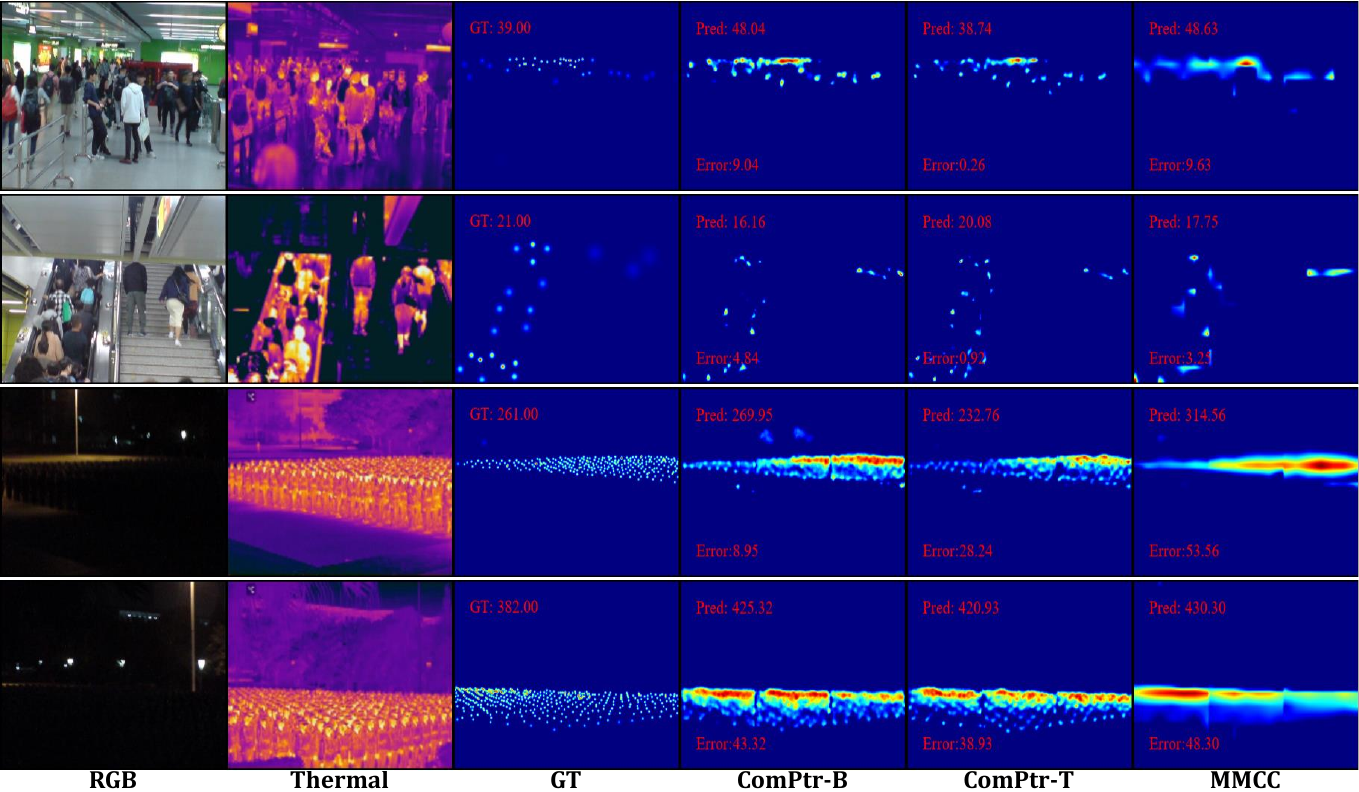}
  \caption{Visual comparison of samples corresponding to the bright scene in the top two rows, and to the low-brightness scene in the bottom two rows for RGB-T crowd counting.
    ``GT'': The number of people annotated.
    ``Pred'': The number estimated by the method.
    ``Error'': The absolute error between ``GT'' and ``Pred''.
  }
  \label{fig:rgbtcc-cmp}
\end{figure*}

\begin{figure*}[!t]
  \centering
  \includegraphics[width=0.7\linewidth]{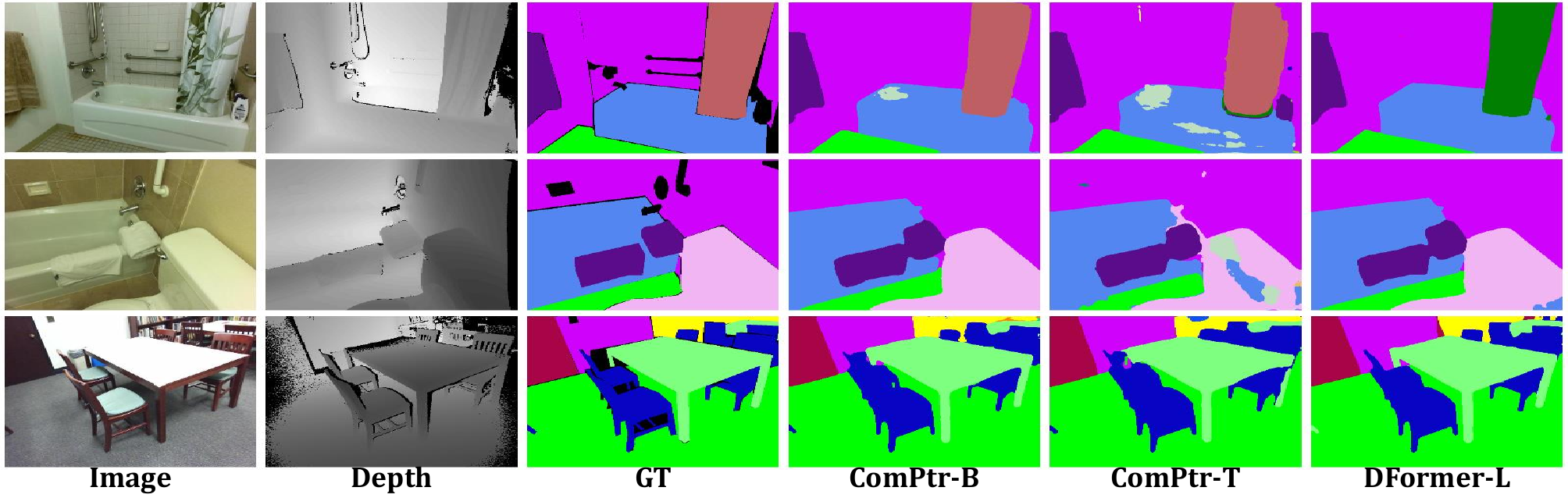}
  \caption{Visual comparison for RGB-D semantic segmentation.}
  \label{fig:mmseg-cmp}
\end{figure*}

\subsection{RGB-D Semantic Segmentation (RGB-D SS)}\label{sec:mmseg-setting}

\noindent\textbf{Datasets.}
We also conduct extended experiments and analysis on the popular indoor \textbf{RGB-D SS} benchmark,
SUN-RGBD dataset with 37 classes~\cite{SUN-RGBD-37classes}.
It contains 10335 pairs of RGB-D images with the shape of $530 \times 730$, which is captured in the indoor scene, and there are 5285 image pairs for training and 5050 image pairs for testing.
It is a representative and challenging RGB-D semantic segmentation dataset and includes complex and diverse indoor categories such as scene structures like walls and floors, furniture items like beds and chairs, and objects like lamps and bags.
Following the practices and strategies of the existing works~\cite{mmseg-DPFormer,mmseg-CMX,mmseg-ShapeConv,mmseg-SA-Gate}, we employ some general data augmentation strategies as~\cite{mmseg-SA-Gate,mmseg-SGNet}, including random flipping, scaling and cropping, and set the number of epochs as 400, batch size as 14, initial learning rate as 0.00001, weight decay as 0.0001, and input size as $480 \times 480$, respectively.
The poly learning rate scheduler~\cite{poly} with a warm-up stage and a factor of 0.9 is imposed on the AdamW optimizer~\cite{AdamW}.

\noindent\textbf{Metrics.}
For RGB-D SS, the commonly used metric \textit{mean region intersection over union} (mIoU) is reported here, and it is defined as follows: $\mathrm{mIoU}   = \frac{1}{N_{cls}} \frac{\sum_i N_{i \rightarrow i}}{\sum_k N_{i \rightarrow k} + \sum_k N_{k \rightarrow i} - N_{i \rightarrow i}}$,
where $N_{i \rightarrow j}$ represents the number of pixels of class $i$ classified as class $j$ by the model and $N_{cls}$ is the number of classes in the dataset.

\subsection{Comparison with State-of-the-Art Methods}
\label{sec:sota-cmp}

We evaluate the competitiveness of our ComPtr by carefully comparing it with the state-of-the-art on different tasks.
There are 16 models and variants from~\cite{changedetection-FCNN,changedetection-FC-EF-Res,changedetection-IFN,changedetection-DTCDSCN,changedetection-SNUNet,changedetection-BIT,changedetection-EGRCNN,changedetection-MSPSNet,changedetection-ICIF-Net,changedetection-ChangeFormer,changedetection-SACNet,changedetection-FTN,changedetection-TransYNet,changedetection-BTNIFormer,changedetection-TTP,changedetection-DMINet} for RSCD,
12 methods for RGB-T CC which contains 6 single-modal methods~\cite{cc-CSRNet,cc-BL,cc-dm,cc-P2PNet,cc-MARUNet-CFANet,cc-MAN} retrained by~\cite{rgbtcc-RGBTMMCC} and 6 bi-modal methods~\cite{rgbtcc-CMCRL-RGBTCC,rgbtcc-TAFNet,rgbtcc-MAT,rgbtcc-DEFNet,rgbtcc-RGBTMMCC,rgbtcc-CCANet},
16 methods~\cite{JLDCF-RGBDSOD-journal,HDFNet,SIP,RD3D-RGBDSOD,TriTransNet-RGBDSOD,DCF-RGBDSOD,DSA2F-RGBDSOD,COME15K-CMINet,SPNet-RGBDSOD-journal,rgbdsod-VST,HAINet-RGBDSOD,CCAFNet-RGBDSOD,DCMF-RGBDSOD,rgbdtsod-SwinNet,caver-tip,rgbdtsod-HRTransNet} for RGB-D SOD,
11 methods~\cite{VT821-MTMR,M3S-NIR-RGBTSOD,VT5000-ADF,VT1000-SDGL,CGFNet-RGBTSOD,CSRNet-RGBTSOD,ECFFNet-RGBTSOD,MIDD-RGBTSOD,rgbdtsod-SwinNet,caver-tip,rgbdtsod-HRTransNet} for RGB-T SOD,
and 19 methods~\cite{mmseg-3DGNN,mmseg-RDFNet,mmseg-ACNet,mmseg-CEN,mmseg-SA-Gate,mmseg-SGNet,mmseg-NANet,mmseg-ShapeConv,mmseg-FSFNet,mmseg-FRNet,mmseg-EMSANet,mmseg-TokenFusion,mmseg-MultiMAE,mmseg-CMNeXt,mmseg-PGDENet,mmseg-CMX,mmseg-EMSAFormer,mmseg-Asymformer,mmseg-DFormer} for RGB-D SS.
In addition, to evaluate the task transfer performance of existing transformer-based methods, we select two algorithms~\cite{rgbdtsod-SwinNet,mmseg-DFormer} with good performance and publicly available code.
By retraining them, evaluations on all data benchmarks can be obtained.

\subsubsection{Quantitative Comparison}

All comparisons are listed in Tab.~\ref{tab:rscd-levir}, Tab.~\ref{tab:rscd-sysu}, Tab.~\ref{tab:rgbtcc}, Tab.~\ref{tab:rgbdsod}, Tab.~\ref{tab:rgbtsod}, and Tab.~\ref{tab:rgbdss}.
From the comparison in Tab.~\ref{tab:rscd-levir} and Tab.~\ref{tab:rscd-sysu}, our method does not obtain the best results on Pre. and Rec., but achieves significant leading performance on F1 and IoU in the RSCD task.
Our ``-B'' version achieves a significant improvement of 0.60 F1 and 0.79 IoU over SAM~\cite{mmseg-SAM}-based TTP~\cite{changedetection-TTP} on LEVIR-CD, and 0.59 F1 and 0.86 IoU over TransY-Net~\cite{changedetection-TransYNet} on SYSU-CD.
In RGB-T CC, ComPtr-T achieves an improvement of 0.38 GAME$_0$ and 0.31 RMSE over the recent~\cite{rgbtcc-RGBTMMCC} as shown in Tab.~\ref{tab:rgbtcc}.
With respect to three bi-modal SOD methods, CAVER~\cite{caver-tip}, SwinNet~\cite{rgbdtsod-SwinNet}, and HRTransNet~\cite{rgbdtsod-HRTransNet}, our performance is still superior, especially on DUTLF-Depth, SIP and STEREO1000 for RGB-D SOD in Tab.~\ref{tab:rgbdsod}, and VT821, VT1000 and VT5000 for RGB-T SOD in Tab.~\ref{tab:rgbtsod}.
Besides, for more complex multi-class RGB-D SS in Tab.~\ref{tab:rgbdss}, our ComPtr-B achieves the best 52.8 mean IoU on the SUN-RGBD dataset.
The two versions, ``-T'' and ``-B '', perform differently on different tasks.
\textbf{\textit{There may be two underlying factors.
    On the one hand, it may be related to the different requirements of hyper-parameters during training for backbone with different volumes~\cite{dl-ScaleModel}.
    On the other hand, the different characteristics of these datasets result in different fitness for the swin architecture.}}
Actually, to avoid over-engineering, we do not validate more parameter combinations and structural variants.
And current results have demonstrated the generality of the proposed architecture and its adaptability to diverse task concepts, data forms, and object attributes.

\subsubsection{Qualitative Comparison}

In Fig.~\ref{fig:rscd-cmp}, Fig.~\ref{fig:rgbtcc-cmp}, Fig.~\ref{fig:rgbdsod-cmp}, Fig.~\ref{fig:rgbtsod-cmp}, and Fig.~\ref{fig:mmseg-cmp}, we visualize the prediction maps of the proposed model and several state-of-the-art methods.
In addition to the results of our method, those of some recent methods with publicly available weight parameters or predictions are also listed.
Visually, the results of our algorithm are much closer to the ground truth, and also adapt well to the complex examples shown here.
For example, in the face of dense small buildings and independent large buildings in Fig.~\ref{fig:rscd-cmp}, and strong background interference in Fig.~\ref{fig:rgbdsod-cmp} and Fig.~\ref{fig:rgbtsod-cmp}, our algorithm performs more robustly.
For the prediction results of the RGB-T CC task in Fig.~\ref{fig:rgbtcc-cmp}, ours has a smaller absolute error compared to MMCC~\cite{rgbtcc-RGBTMMCC}'s.
It is also worth noting that our method still shows good crowd counting performance even in low-light scenarios due to the assistance of thermal infrared images.
Besides, as shown in Fig.~\ref{fig:mmseg-cmp} for RGB-D SS, the predictions of our approaches also exhibit better intra-class consistency and inter-class differentiation.
These effects can be attributed to the generality of the concept of complementarity for these tasks and the effectiveness of component design based on consistency and difference.

\subsection{Ablation Studies}

A series of ablation studies are presented in this section to reveal how the proposed components affect the final performance of the model.
These experiments are constructed on three representative tasks, RGB-T CC, RSCD and RGB-D SOD, with large-scale datasets of various forms and attributes, where we take Swin-T~\cite{Swin} as the backbone.

\subsubsection{Number of Proxy Prototypes}
\label{sec:num_k}

\begin{table}[!t]
  \centering
  \resizebox{0.7\linewidth}{!}{%
    \rowcolors{2}{gray!10}{white}
    \begin{tabular}{l|*{1}{c}|*{1}{c}|*{3}{c}|c}
  \toprule[2pt]
                             &
  \multicolumn{1}{|c}{RGBT-CC~\cite{rgbtcc-CMCRL-RGBTCC}}
                             &
  \multicolumn{1}{|c}{LEVIR-CD~\cite{changedetection-LEVIR}}
                             &
  \multicolumn{3}{|c|}{STEREO1000~\cite{STEREO}}
                             &                                                                                                                                             \\
  \multirow{-2}{*}{Variants} & GAME$_0$~$\downarrow$ & IoU~$\uparrow$ & $S_m$~$\uparrow$ & $F^{\omega}_{\beta}$~$\uparrow$ & $E_m$~$\uparrow$ & \multirow{-2}{*}{$\Delta$} \\
  \midrule[1pt]
  ADA ($K=1$)                & 11.59                 & 86.11          & 91.94            & 87.41                           & 95.42            & +0.00\%                    \\
  ADA ($K=2$)                & 11.55                 & 86.20          & 91.95            & 87.44                           & 95.45            & +0.11\%                    \\
  ADA ($K=4$)                & \best{10.52}          & \best{86.25}   & \best{91.99}     & \best{87.45}                    & \best{95.47}     & +1.91\%                    \\
  ADA ($K=8$)                & 11.53                 & 86.09          & 91.92            & 87.15                           & 95.40            & +0.03\%                    \\
  ADA ($K=16$)               & 11.00                 & 86.06          & 91.89            & 87.38                           & 95.32            & +0.97\%                    \\
  \midrule[1pt]
  ADA ($K=\infty$)           & 10.84                 & 86.01          & 91.96            & 87.43                           & 95.44            & +1.28\%                    \\
  Std.                       & 11.11                 & 86.05          & 91.72            & 87.26                           & 95.41            & +0.73\%                    \\
  \bottomrule[2pt]
\end{tabular}

  }
  \caption{Ablation on the number $K$ of the proxy prototypes in each attention layer.
    ``$K=4$'' is our final choice.
    ``$K=\infty$'': With the same number of proxy prototypes as that of the input tokens.
    ``Std.'': Dense interactive form of standard attention.
    ``$\Delta$'': Relative performance gain.
  }
  \label{tab:num_k}
\end{table}

The number of proxy prototypes is an important hyper-parameter of the proposed ADA.
To simplify the design process, the current version of the model uses the same number of prototypes in both types of embedded components.
We set five small constant values $\{1, 2, 4, 8, 16\}$ for $K$ as candidates and record the performance of the five variants in Tab.~\ref{tab:num_k} on three different tasks, RGB-T CC, RSCD and RGB-D SOD, respectively.
As we can see from the results, the sensitivity to $K$ varies across tasks.
\textit{RSCD and RGB-D SOD tasks are relatively more robust to the value of $K$.}
Considering the balance of performance across tasks, we choose $K=4$ as the final setting.
Besides, we also supplement two special cases, namely ``$K=\infty$'' where the value of $K$ is consistent with the number of input tokens and ``Std.'' where the proposed ADA is replaced with the full standard attention that also integrates CompOps for a reasonable comparison.
Notably, the version $K=4$ surpasses the two benchmark versions, reflecting the potential of more compact representations for consistency and difference modelling.
From the complexity statistics of different variants in Fig.~\ref{fig:efficiency}, we can see that the proposed strategy based on the information aggregation mediated by a small number of prototypes can effectively save the computational budget of the inference process, which actually holds true for the training phase as well, and improve the parallel efficiency of the model.

\begin{table}[!t]
  \centering
  \resizebox{\linewidth}{!}{%
    \rowcolors{2}{gray!10}{white}
    \begin{tabular}{l|rrr|*{1}{c}|*{1}{c}|*{3}{c}|c}
  \toprule[2pt]
                             &
  FLOPs
                             &
  Params.
                             &
  FPS
                             &
  \multicolumn{1}{c}{RGBT-CC~\cite{rgbtcc-CMCRL-RGBTCC}}
                             &
  \multicolumn{1}{|c}{LEVIR-CD~\cite{changedetection-LEVIR}}
                             &
  \multicolumn{3}{|c|}{STEREO1000~\cite{STEREO}}
                             &                                                                                                                                                                  \\
  \multirow{-2}{*}{Variants} & (G)  & (M)  &      & GAME$_0$~$\downarrow$ & IoU~$\uparrow$ & $S_m$~$\uparrow$ & $F^{\omega}_{\beta}$~$\uparrow$ & $E_m$~$\uparrow$ & \multirow{-2}{*}{$\Delta$} \\
  \midrule[1pt]
  ComPtr                     & 33.9 & 38.3 & 30.8 & \best{10.52}          & \best{86.25}   & \best{91.99}     & \best{87.45}                    & \best{95.47}     & -0.00\%                    \\
  \quad Swap Both            & 33.9 & 38.3 & 30.8 & 12.03                 & 82.65          & 90.11            & 84.79                           & 94.45            & -4.94\%                    \\
  \quad w/o CEB              & 31.2 & 31.2 & 36.0 & 11.14                 & 85.41          & 91.26            & 86.28                           & 94.75            & -1.95\%                    \\
  \quad w/o DAB              & 32.6 & 34.8 & 38.1 & 11.22                 & 84.30          & 90.43            & 85.79                           & 94.53            & -2.70\%                    \\
  \quad w/o Both             & 29.9 & 27.7 & 49.9 & 12.53                 & 81.82          & 89.65            & 83.54                           & 94.03            & -6.55\%                    \\
  \midrule[0.5pt]
  \quad w/o CompOps          & 33.5 & 37.5 & 32.6 & 11.20                 & 84.81          & 91.43            & 86.72                           & 94.79            & -2.06\%                    \\
  \quad w/ CompOps (L)       & 33.5 & 37.5 & 31.3 & 10.97                 & 85.86          & 91.79            & 87.25                           & 95.18            & -1.10\%                    \\
  \quad w/ CompOps (LA)      & 33.7 & 37.9 & 31.0 & 10.88                 & 85.98          & 91.81            & 87.33                           & 95.22            & -0.87\%                    \\
  \quad w/ CompOps (LAA)     & 33.9 & 38.3 & 30.8 & \best{10.52}          & \best{86.25}   & \best{91.99}     & \best{87.45}                    & \best{95.47}     & -0.00\%                    \\
  \quad w/ CompOps (LAAA)    & 34.0 & 38.7 & 29.8 & 11.02                 & 85.21          & 91.93            & 87.28                           & 95.34            & -1.27\%                    \\
  \bottomrule[2pt]
\end{tabular}

  }
  \caption{Ablation on the proposed components in the proposed ComPtr.
    ``w/o CEB'': Without the consistency enhancement block.
    ``w/o DAB'': Without the difference awareness block.
    ``w/o Both'': Baseline model which is a variant without the proposed blocks.
      {``Swap Both'': Swap CEB and DAB.}
    ``w/o CompOps'': ComPtr variant without the complementarity operations.
    ``L'' and ``A'' represent the linear and average pooling layers and ``LAA'' is our default setting.
  }
  \label{tab:ablation}
\end{table}

\begin{figure}[!t]
  \centering
  \includegraphics[width=\linewidth]{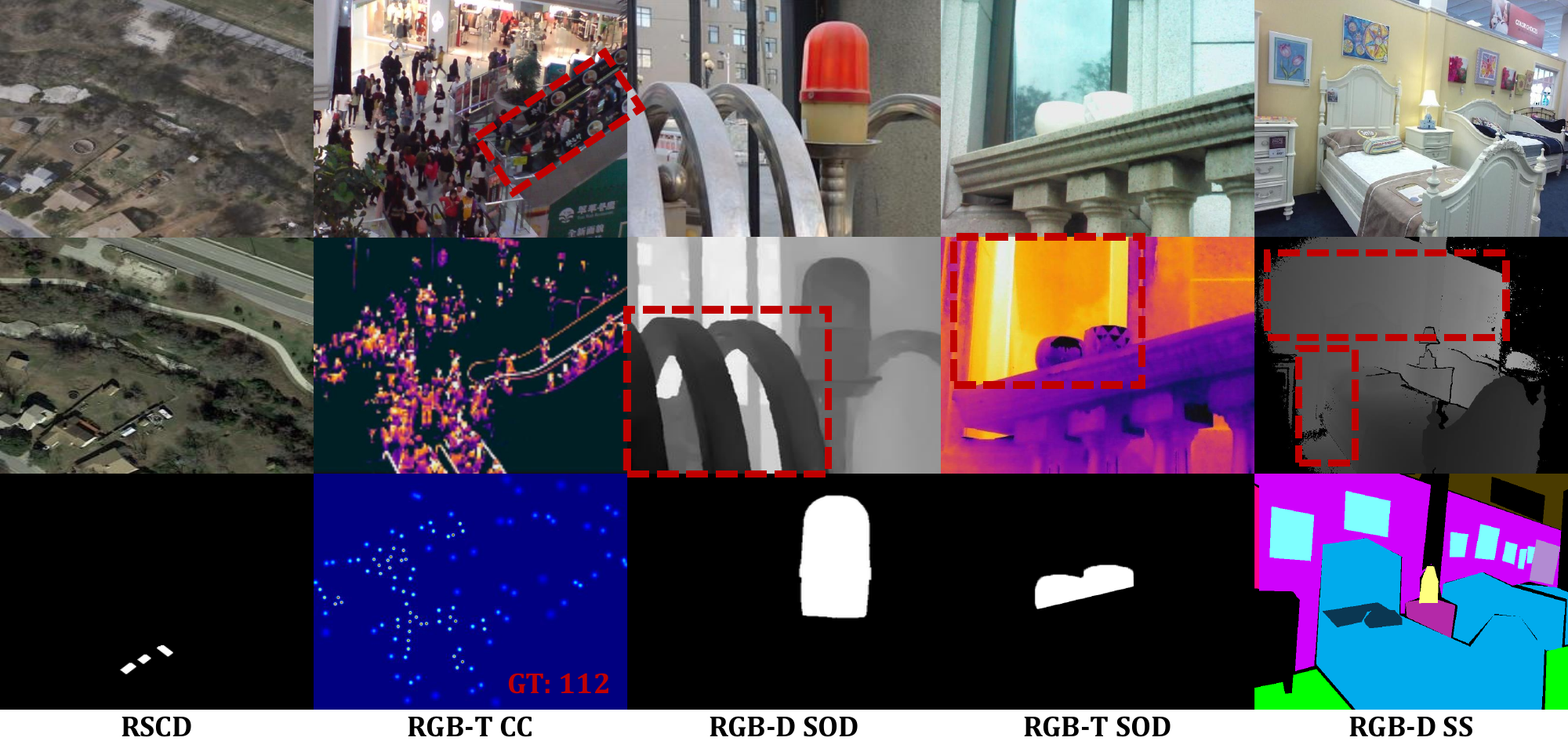}
  \caption{{Representative sample reflecting task challenges.}}
  \label{fig:special_cases}
\end{figure}

\begin{table}[!t]
  \centering
  \resizebox{0.7\linewidth}{!}{%
    \rowcolors{2}{gray!10}{white}
    \begin{tabular}{l|c|c|c|c|c}
    \toprule[2pt]
    Model             & RSCD           & RGBT-CC               & RGB-D SOD      & RGB-T SOD      & RGB-D SS        \\
                      & IoU$~\uparrow$ & GAME$_0$$~\downarrow$ & $S_m~\uparrow$ & $S_m~\uparrow$ & mIoU$~\uparrow$ \\
    \midrule[1pt]
    ComPtr            & 79.44          & 12.33                 & 90.08          & 89.74          & 67.1            \\
    \quad w/o CEB     & 73.63          & 13.79                 & 85.39          & 82.93          & 58.3            \\
    \quad w/o DAB     & 68.98          & 13.20                 & 83.83          & 81.25          & 52.9            \\
    \quad w/o Both    & 65.03          & 14.69                 & 79.82          & 73.19          & 45.1            \\
    \midrule[1pt]
    \quad w/o CompOps & 70.12          & 13.47                 & 85.78          & 83.16          & 56.5            \\
    \bottomrule[2pt]
\end{tabular}

  }
  \caption{{Ablation on the task-specific test subsets.}}
  \label{tab:ablation_challenges}
\end{table}

\subsubsection{Effectiveness of Proposed Components}

To verify the effectiveness of the proposed components, we remove all complementarity blocks, CEBs and DABs, respectively, and evaluate these variants on the three datasets from different tasks, RGB-T CC, RSCD, and RGB-D SOD.
The results in Tab.~\ref{tab:ablation} demonstrate that these components provide a positive gain for the performance of the final model.
Besides, there is an interesting observation that removing the difference component causes significant performance degradation (-2.70\%).
This highlights the importance of difference information for bi-source tasks.
At the same time, the consistency component also reflects the positive effect.
In addition, if an attempt is made to remove the complementarity operations in ComPtr and only apply a pure cross-source transformer architecture, the model performance will degrade significantly (-2.06\%), as shown in ``w/o CompOps'' of Tab.~\ref{tab:ablation}.
{When we swap the positions of CEB and DAB, our experiments reveal significant performance degradation, particularly on the difference-sensitive RSCD task (LEVIR-CD dataset), as shown in ``Swap Both'' of Tab.~\ref{tab:ablation}.
This suggests that consistency enhancement in encoder representations improves common object representations, while difference awareness facilitates decoder feature refinement.
Swapping these operations introduces functional misalignment with the inherent roles of encoder-decoder architectures in dense prediction tasks.}
We also analyze the hierarchical structure in complementary operations.
To avoid over-engineering, we use similar structures in both complementarity operations.
As seen from the results in Tab.~\ref{tab:ablation}, the default setting ``LAA'' has better performance.

  {Each task presents diverse modeling challenges as shown in Fig.~\ref{fig:special_cases}, including the foreground-background imbalance in RSCD, the crowd occlusion in RGB-T CC, the modality interference in RGB-D/T SOD, and the ineffective depth reference in RGB-D SS.
    To investigate our method's advantages to address these challenges, we conduct targeted experiments by sampling 300 representative test samples for each task and the results are summarized in Tab.~\ref{tab:ablation_challenges}.
    In the table, the proposed modules consistently demonstrate positive performance gains in these tasks by effectively addressing specific challenges.
    These advantages can be attributed to the proposed algorithm's ability to model global contextual information effectively, as well as its strong adaptability with decoupled complementarity representation learning.}

\begin{table}[!t]
  \centering
  \resizebox{\linewidth}{!}{%
    \rowcolors{2}{gray!10}{white}
    \begin{tabular}{l|rrr|*{1}{c}|*{1}{c}|*{3}{c}|c}
  \toprule[2pt]
                             &
  FLOPs
                             &
  Params.
                             &
  FPS
                             &
  \multicolumn{1}{c}{RGBT-CC~\cite{rgbtcc-CMCRL-RGBTCC}}
                             &
  \multicolumn{1}{|c}{LEVIR-CD~\cite{changedetection-LEVIR}}
                             &
  \multicolumn{3}{|c|}{STEREO1000~\cite{STEREO}}
                             &                                                                                                                                                                  \\
  \multirow{-2}{*}{Variants} & (G)  & (M)  &      & GAME$_0$~$\downarrow$ & IoU~$\uparrow$ & $S_m$~$\uparrow$ & $F^{\omega}_{\beta}$~$\uparrow$ & $E_m$~$\uparrow$ & \multirow{-2}{*}{$\Delta$} \\
  \midrule[1pt]
  1,1,1,1 (ComPtr-T)         & 33.9 & 38.3 & 30.8 & 10.52                 & 86.25          & 91.99            & 87.45                           & 95.47            & -0.00\%                    \\
  1,2,3,4                    & 35.4 & 47.5 & 21.4 & 10.92                 & 86.13          & 92.07            & 88.14                           & 95.59            & -0.59\%                    \\
  2,2,6,2 (Same as Swin-T)   & 36.3 & 44.6 & 19.6 & 10.78                 & 86.03          & 91.95            & 87.51                           & 95.32            & -0.57\%                    \\
  2,2,2,2                    & 35.3 & 41.8 & 24.6 & 10.61                 & 86.18          & 91.29            & 87.53                           & 94.98            & -0.42\%                    \\
  4,4,4,4                    & 38.1 & 48.9 & 18.1 & 11.01                 & 86.09          & 92.17            & 88.36                           & 95.50            & -0.72\%                    \\
  \bottomrule[2pt]
\end{tabular}

  }
  \caption{{Ablation on the depth settings of the decoder.
        In each variant, the four numbers represent the number of DEBs in each stage, from shallow to deep layers.}}
  \label{tab:ablation_depth}
\end{table}

\begin{table}[!t]
  \centering
  \resizebox{\linewidth}{!}{%
    \rowcolors{2}{gray!10}{white}
    \begin{tabular}{l|rrr|*{1}{c}|*{1}{c}|*{3}{c}|c}
  \toprule[2pt]
                             &
  FLOPs
                             &
  Params.
                             &
  FPS
                             &
  \multicolumn{1}{c}{RGBT-CC~\cite{rgbtcc-CMCRL-RGBTCC}}
                             &
  \multicolumn{1}{|c}{LEVIR-CD~\cite{changedetection-LEVIR}}
                             &
  \multicolumn{3}{|c|}{STEREO1000~\cite{STEREO}}
                             &                                                                                                                                                                  \\
  \multirow{-2}{*}{Variants} & (G)  & (M)  &      & GAME$_0$~$\downarrow$ & IoU~$\uparrow$ & $S_m$~$\uparrow$ & $F^{\omega}_{\beta}$~$\uparrow$ & $E_m$~$\uparrow$ & \multirow{-2}{*}{$\Delta$} \\
  \midrule[1pt]
  Baseline                   & 29.9 & 27.7 & 49.9 & 12.53                 & 81.82          & 89.65            & 83.54                           & 94.03            & +0.00\%                    \\
  \midrule[0.25pt]
  1 stage w/ DAB             & 30.1 & 30.2 & 44.6 & 12.30                 & 82.28          & 90.11            & 84.24                           & 94.37            & +0.82\%                    \\
  2 stages w/ DAB            & 30.3 & 30.9 & 40.0 & 11.93                 & 83.31          & 90.42            & 84.96                           & 94.62            & +1.96\%                    \\
  3 stages w/ DAB            & 30.6 & 31.1 & 38.0 & 11.49                 & 85.09          & 91.08            & 85.93                           & 94.79            & +3.51\%                    \\
  4 stages w/ DAB            & 31.2 & 31.2 & 34.9 & 11.14                 & 85.41          & 91.26            & 86.28                           & 94.75            & +4.26\%                    \\
  \midrule[0.25pt]
  1 stage w/ CEB             & 30.5 & 33.1 & 46.0 & 12.56                 & 81.99          & 89.75            & 84.09                           & 94.21            & +0.19\%                    \\
  2 stages w/ CEB            & 31.2 & 34.4 & 42.4 & 12.33                 & 83.25          & 90.03            & 84.82                           & 94.26            & +1.11\%                    \\
  3 stages w/ CEB            & 31.9 & 34.7 & 39.9 & 11.82                 & 84.04          & 90.46            & 85.69                           & 94.43            & +2.46\%                    \\
  4 stages w/ CEB            & 32.6 & 34.8 & 35.4 & 11.22                 & 84.30          & 90.43            & 85.79                           & 94.53            & +3.52\%                    \\
  \bottomrule[2pt]
\end{tabular}

  }
  \caption{{Ablation on the number of stages with DAB and CEB.
        The modules are progressively integrated from deeper to shallower stages.}}
  \label{tab:ablation_num_ceb_dab}
\end{table}

\begin{figure*}[!t]
  \centering
  \subfloat[RSTD samples.]{\includegraphics[width=0.33\textwidth]{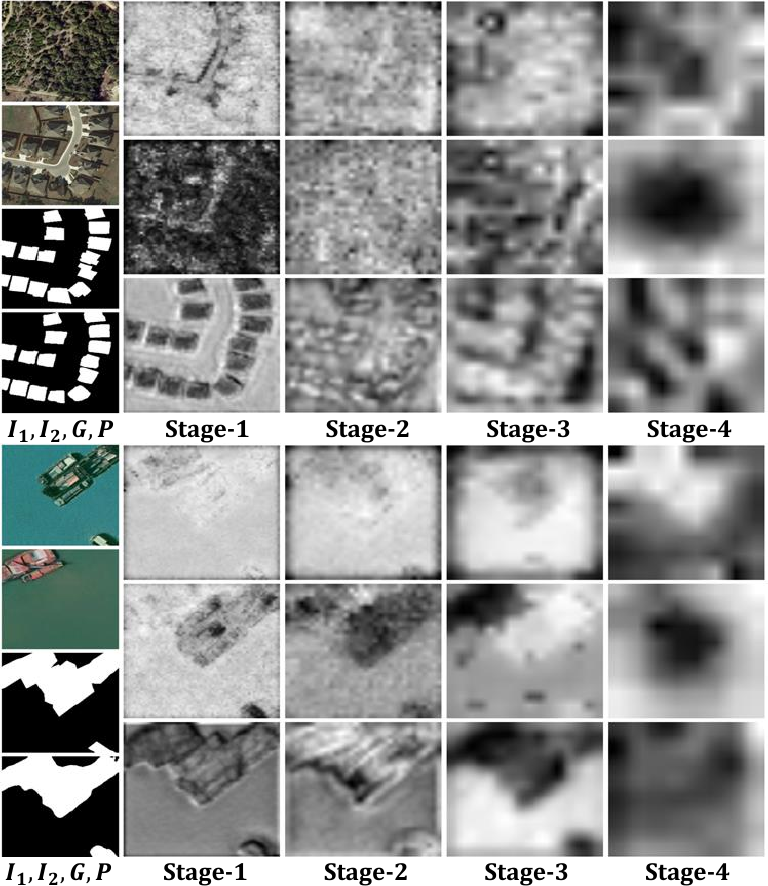}}
  \subfloat[RGB-D SOD samples.]{\includegraphics[width=0.33\textwidth]{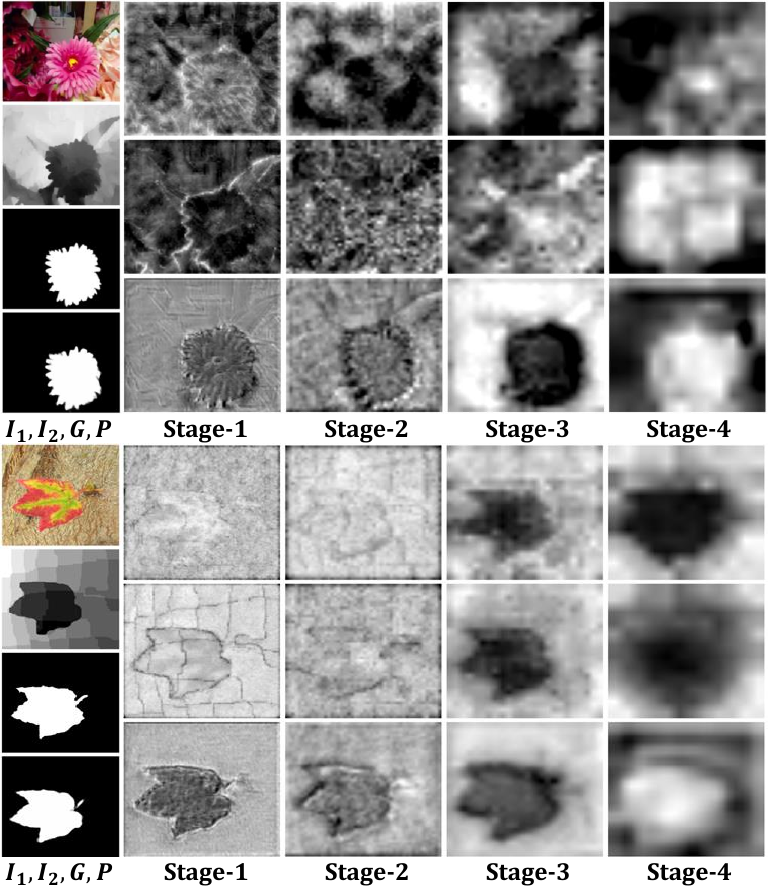}}
  \subfloat[RGB-T SOD samples.]{\includegraphics[width=0.33\textwidth]{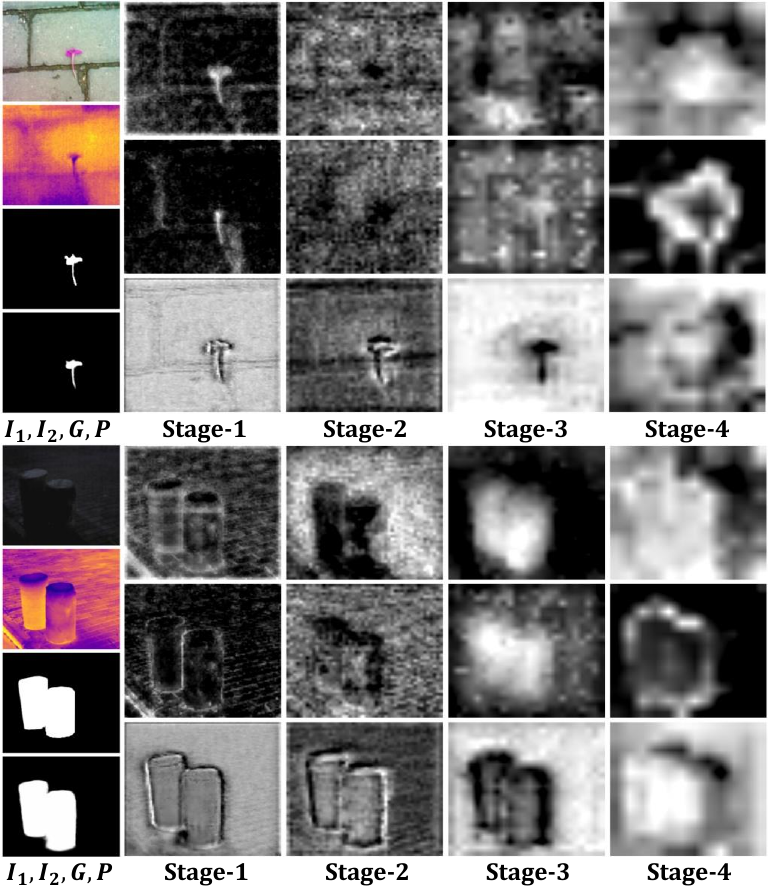}}
  \caption{{Source feature maps in consistency enhancement and difference awareness blocks from different stages.
        These maps are obtained using bilinear interpolation after being averaged along the channels.
        The columns ``Stage-$\star$'' correspond to the consistency and difference source feature maps, and the output feature maps from the corresponding stage, respectively.}
  }
  \label{fig:feature}
\end{figure*}

\begin{figure*}[!t]
  \centering
  \includegraphics[width=0.8\textwidth]{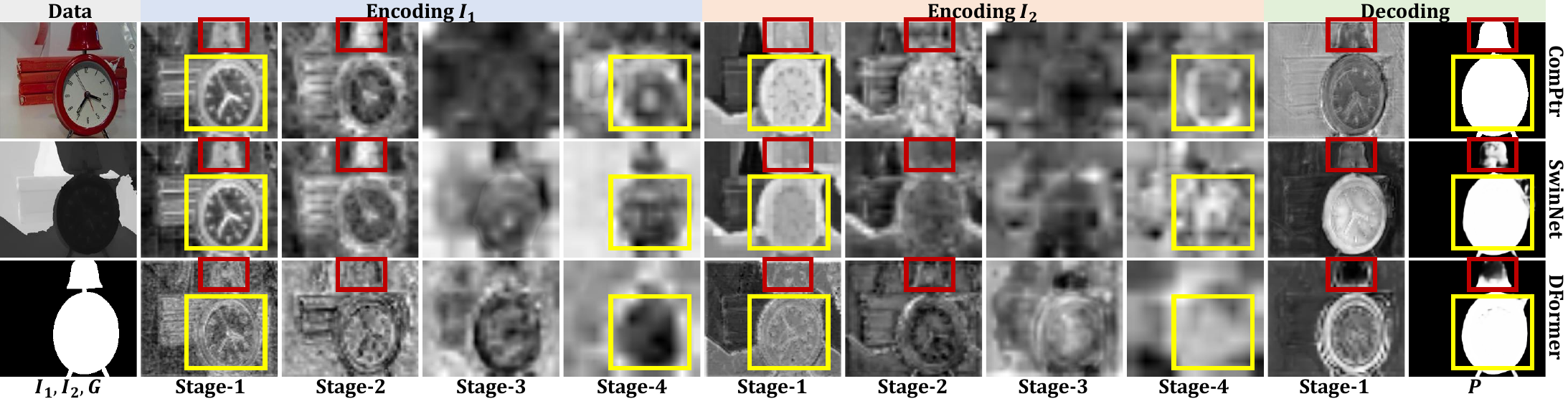}
  \caption{{Visualization of the features from the encoder and decoder of different methods.}}
  \label{fig:feature-supp}
\end{figure*}

\subsubsection{{Decoder Depth Analysis}}

{To avoid over-engineering, we adopt a single DEB for each stage, a setting similar to many existing works~\cite{rgbdtsod-SwinNet,caver-tip,changedetection-DMINet}.
  For further analysis, we perform an ablation study on the decoder depth settings as shown in Tab.~\ref{tab:ablation_depth}.
  As shown in the results, directly adjusting the decoder depth does not result in a positive improvement in overall performance.
  This also implies that under the current simple configuration (``1,1,1,1''), the proposed method is sufficient to function effectively.}

\subsubsection{{Number of Stages with DAB and CEB}}

{To investigate the impact of DAB and CEB on model performance, we conduct experiments based on the baseline (``w/o Both'') in Tab.~\ref{tab:ablation}.
  We gradually introduce CEB and DAB from deep to shallow layers on the baseline to enhance the representation at different stages of the model.
  The experimental results in Tab.~\ref{tab:ablation_num_ceb_dab} show that the model performance is steadily improved with the introduction of CEB and DAB, which also implies that the proposed components play an important role in feature modeling at different stages of the model.}

\subsubsection{{Efficiency of Proposed Modules}}

{%
  We analyze the efficiency of the proposed modules in terms of computational load (FLOPs), inference time (FPS), and the number of parameters (Params.) based on the input size of $256 \times 256$, following the settings in Fig.~\ref{fig:efficiency}.
  All results are presented in Tab.~\ref{tab:ablation}, Tab.~\ref{tab:ablation_depth}, and Tab.~\ref{tab:ablation_num_ceb_dab}.
  And they confirm that each component contributes notable performance gains while adding only marginal computational costs.
  This lightweight yet effective design philosophy ensures the balance between effectiveness and efficiency through carefully crafted feature integration.
}

\begin{figure}[!t]
  \centering
  \subfloat[Cosine Similarity]{\label{fig:similarity}
    \centering
    \includegraphics[width=0.765\linewidth]{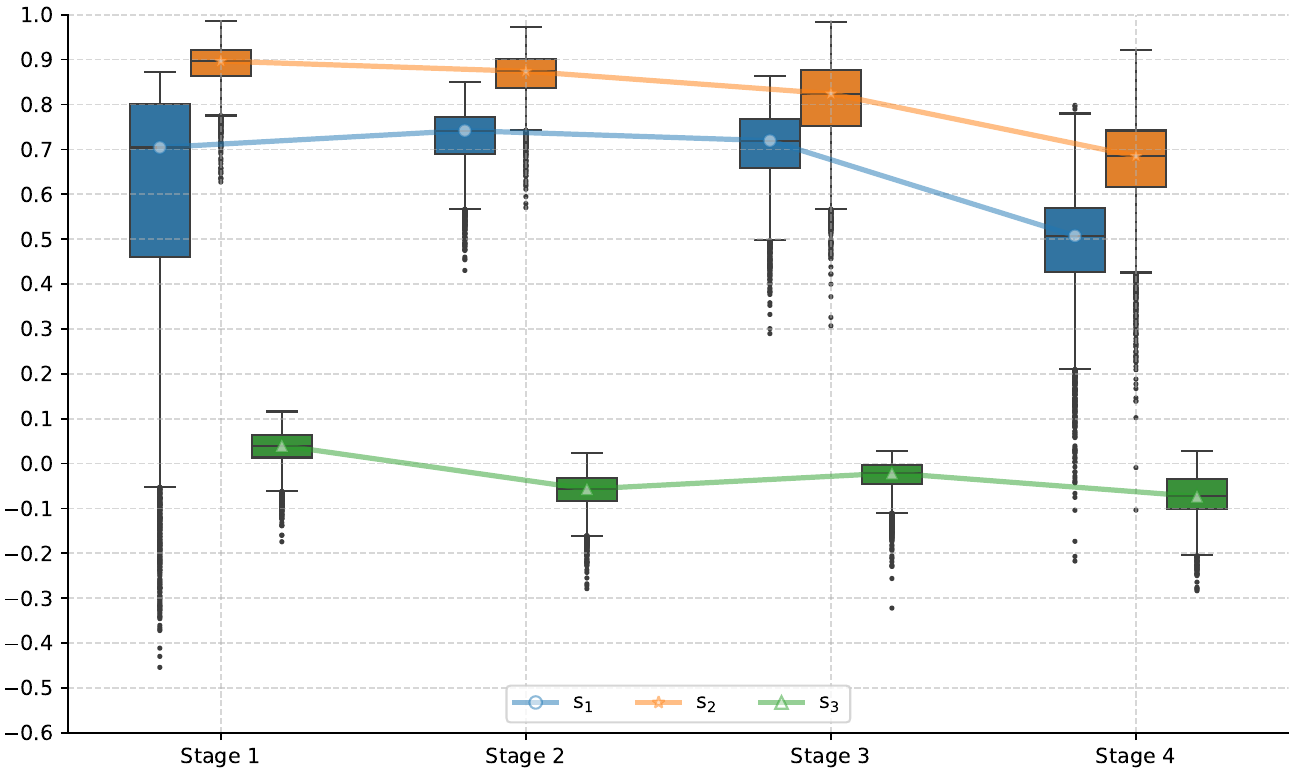}
  }
  \subfloat[Features]{\label{fig:ceb_dab_sim}
    \centering
    \includegraphics[width=0.225\linewidth]{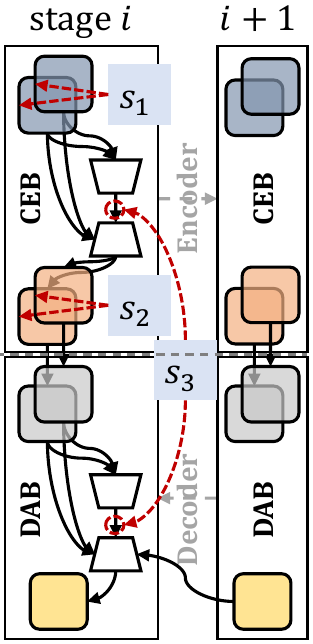}
  }
  \caption{{Feature similarity analysis.}}
\end{figure}

\subsubsection{Feature Analysis}

To better understand the behavior of the proposed components, the analysis is carried out in dense multi-object {scenarios} and sparse single-object {scenarios} with various scales {and shapes}.
And the corresponding source features are visualized in Fig.~\ref{fig:feature}.
It is clear that the two operations play complementary roles in different stages.
In the shallow layer (Stage-1 and 2), the detailed information from the difference operation and the object cues from the consistency operation are paid special attention.
In the deep layer (Stage-3 and 4), the two operations show task-specific attributes.
They converge to the similar pattern and the location of the object of interest in the RGB-D SOD task, while the difference operation provides the model with important cues for the perception of the target area in RSCD.
  {
    Besides, Fig.~\ref{fig:feature-supp} visualizes intermediate features from each encoder stage and the final decoder output for fair comparison.
    Our consistency-enhanced encoder retains clearer object structures and details (notably in Stage-1/2/4), while SwinNet and DFormer produce blurred or fragmented features.
    Additionally, our difference-aware decoder shows more precise and complete object highlighting, proving its effectiveness in enhancing critical visual representations.
  }

  {Further, we analyze the feature cosine similarity to verify the modeling of consistency and difference, as shown in Fig.~\ref{fig:similarity}.
    For the CEB, the overall inter-stream semantic similarity declines from Stage 1 to Stage 4, reflecting the emergence of stream-specific semantics.
    However, the CEB's output similarity $s_2$ remains higher than its input $s_1$ at each stage, signifying that it strengthens common object representations.
    We also compute the similarity $s_3$ between outputs from the aggregation stages within the CEB and the DAB.
    These outputs inherently aggregate attribute-specific representations derived from the complementarity operation (consistency and difference computation).
    The results show they consistently maintain very low similarity (close to 0) across all stages.
    This nearly orthogonal relationship with the consistency representation visually demonstrates the effectiveness of modeling difference cues.
    These stage-wise patterns succinctly confirm the modeling characteristics of these modules.}

\begin{table}[!t]
  \centering
  \resizebox{0.9\linewidth}{!}{\begin{tabular}{l|c|c|c|c|c|c}
    \toprule[2pt]
    \multirow{2}{*}{Variants}                        & LEVIR-CD       & RGBT-CC             & SUN-RGBD        & STEREO1000    & VT5000        & \multirow{2}{*}{$\Delta$} \\
                                                     & IoU$~\uparrow$ & GAME$_0~\downarrow$ & mIoU$~\uparrow$ & $S_m~\uparrow$ & $S_m~\uparrow$ &                           \\
    \midrule[1pt]
    ComPtr-T                                         & 86.25          & 10.52               & 48.9            & 91.99         & 90.71         & +0.00\%                   \\
    Joint Learning                                   & 86.33          & 10.49               & 49.5            & 92.84         & 91.95         & +0.78\%                   \\
    \bottomrule[2pt]
\end{tabular}
}
  \caption{{Comparison of different training settings.}}
  \label{tab:joint_learning}
\end{table}

\subsubsection{{Multi-Task Joint Learning}}

{%
  In this work, we position the proposed method as a transferable general single-task architecture.
  \textbf{This is to facilitate fair comparisons with task-specific expert models based on the same training setup, thereby eliminating the potential impact of data differences.}
  But our architecture can be readily extended to a multi-task joint modeling form through simple modifications.
  The model can integrate multiple task-specific predictors.
  They share the same consistency-enhanced encoder and difference-aware decoder representations.
  During training, we {employ} the cyclic task sampling strategy with synchronized gradient updates.
  After each full cycle of multi-task exposure, we perform collective back-propagation to balance learning across tasks.
  As shown in Tab.~\ref{tab:joint_learning}, the joint learning with multi-task data contributes positively to the overall performance of the model.
}

\begin{figure}[!t]
  \centering
  \includegraphics[width=\linewidth]{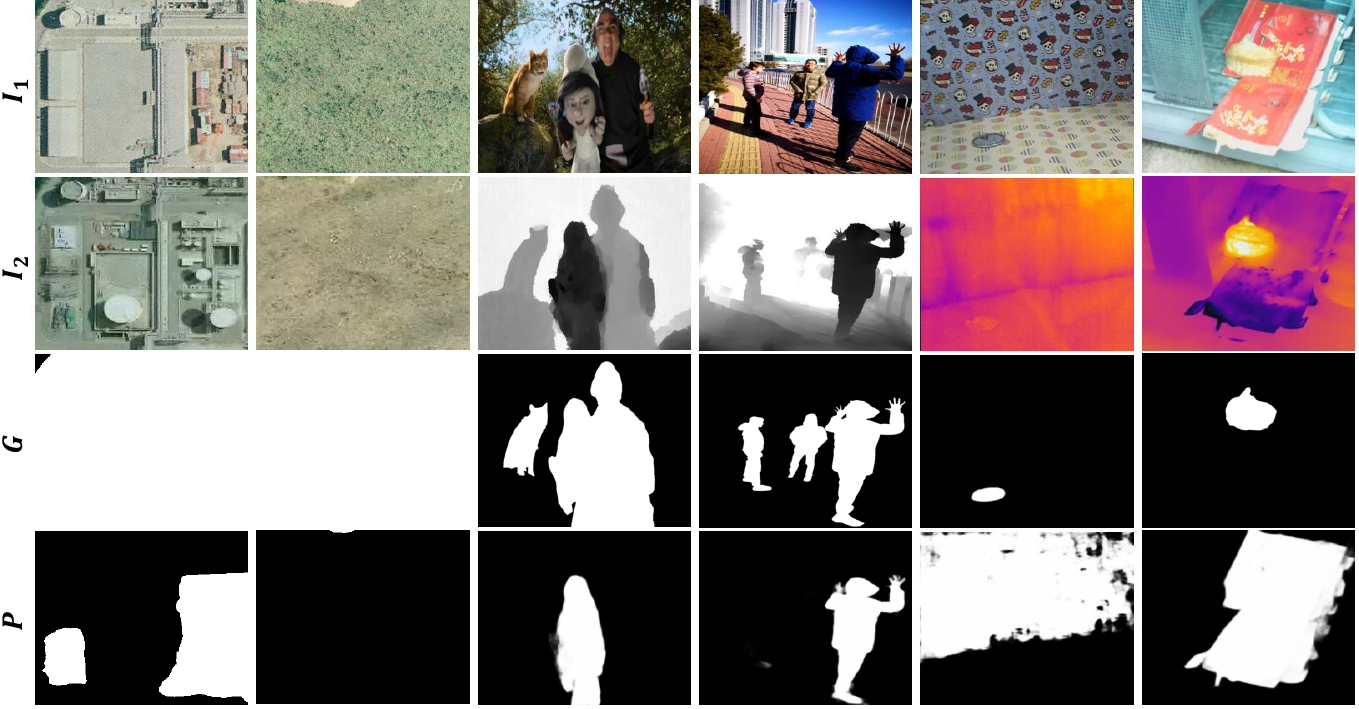}
  \caption{Some typical failure cases.}
  \label{fig:failedcase}
\end{figure}

\subsubsection{Typical Failure Cases}

We list some typical failure cases in Fig.~\ref{fig:failedcase}.
These visualizations mainly involve the ambiguity in the definition of objects of interest in the annotation process of existing datasets.
This is introduced by the relative character of these concepts themselves.
In fact, the question of ``\textit{what level of scene change or object saliency is required for a scene to be considered changed or an object to be considered salient}'' still lacks the necessary research in the current community.

\section{Conclusion}

In this paper, we design a simple yet effective task-generic framework, ComPtr, from the concept of complementarity.
It is common in bi-source information interaction for diverse dense prediction tasks with significantly different data patterns.
From the perspective of consistency and difference, we design the consistency enhancement block and the difference awareness block respectively, which meet the needs of feature extraction and prediction decoding for complementary information.
At the same time, we design an optimization strategy mediated by the proxy prototype, which effectively reduces the computational complexity of the global complementary information propagation process.
Based on it, we construct a novel clustering-inspired aggregation-diffusion attention mechanism to improve the flexibility and scalability of the model.
These simple yet effective designs make the proposed model no longer depend on task-specific data properties or forms, achieving a more general architecture.
Extensive experiments on five diverse dense prediction tasks verify the effectiveness of the proposed method with superior performance to existing state-of-the-art competitors.

\ifCLASSOPTIONcaptionsoff
  \newpage
\fi

\bibliographystyle{IEEEtran}
\bibliography{IEEEabrv,bib_lite}

\end{document}